\documentclass[12pt,onecolumn,draftclsnofoot]{IEEEtran}

\usepackage{graphicx}
\usepackage{float}
\usepackage[caption=false]{subfig}
\usepackage[space]{grffile}
\usepackage{latexsym}
\usepackage{enumerate}
\usepackage{textcomp}
\usepackage{amsfonts,amsmath,amssymb}
\usepackage{bm}
\DeclareMathOperator*{\argmax}{\arg\!\max}
\usepackage{url}
\usepackage{algorithm}
\usepackage{algpseudocode}
\usepackage{tabularx}
\usepackage{longtable}
\usepackage{multirow,booktabs,makecell}
\usepackage{xcolor}
\usepackage{diagbox}
\usepackage{siunitx}
\usepackage{soul}
\usepackage{cite}
\usepackage{layouts}
\usepackage{svg}
\usepackage{lengthconvert}
\usepackage[affil-sl]{authblk}

\usepackage{caption}
\newcommand{\BO}{TPBO}

\newcommand{\fy}{y}

\newcommand{\fv}{\bm{\mathrm{f}}}
\newcommand{\fs}{\mathcal{S}}

\newcommand{\fa}{a(\bovar)}
\newcommand{\bovar}{\mathbf{q}}
\newcommand{\bovarspace}{\mathcal{Q}}

\newcommand{\Jnees}[0]{J_{NEES}}
\newcommand{\Jnis}[0]{J_{NIS}}
\newcommand{\Cnees}[0]{C_{NEES}}
\newcommand{\Cnis}[0]{C_{NIS}}
\newcommand{\Vnis}[0]{V_{NIS}}

\def\mr[#1]#2#3{\multirowcell{#2}[#1]{#3}}

\newcommand{\Ignore}[1]{}

\newcommand{\kLst}{k-1}
\newcommand{\kCur}{k}

\newcommand{\x}[1]{\mathbf{x}_{#1}}
\newcommand{\z}[1]{\mathbf{z}_{#1}}

\newcommand{\uVec}[1]{\mathbf{u}_{#1}}
\newcommand{\vVec}[1]{\mathbf{v}_{#1}}
\newcommand{\wVec}[1]{\mathbf{w}_{#1}}
\newcommand{\F}[1]{\mathbf{F}_{#1}}
\newcommand{\Ft}[1]{\mathbf{F}_{#1}^\top}
\newcommand{\HM}[1]{\mathbf{H}_{#1}}
\newcommand{\HMt}[1]{\mathbf{H}_{#1}^\top}
\newcommand{\Q}[1]{\mathbf{Q}_{#1}}
\newcommand{\R}[1]{\mathbf{R}_{#1}}
\newcommand{\ex}[1]{\mathbf{e}_{\mathbf{x},#1}}
\newcommand{\ez}[1]{\mathbf{e}_{\mathbf{z},#1}}
\newcommand{\nees}[1]{\epsilon_{\mathbf{x},#1}}
\newcommand{\nis}[1]{\epsilon_{\mathbf{z},#1}}
\newcommand{\avgnees}[1]{\bar{\epsilon}_{\mathbf{x},#1}}
\newcommand{\avgnis}[1]{\bar{\epsilon}_{\mathbf{z},#1}}
\newcommand{\W}[0]{\mathbf{W}}
\newcommand{\V}[0]{\mathbf{V}}

\newcommand{\E}[1]{\mathbb{E}\left[#1\right]}

\newcommand{\ECondOuter}[2]{\E{{#1}{#1}^\top|#2}}
\newcommand{\xCond}[2]{\hat{\mathbf{x}}_{#1|#2}}

\newcommand{\covCond}[2]{\mathbf{P}_{#1|#2}}

\newcommand{\zCond}[2]{\hat{\mathbf{z}}_{#1|#2}}

\newcommand{\innovCov}[2]{\mathbf{S}_{#1|#2}}
\newcommand{\nx}[0]{n_{\mathbf{x}}}
\newcommand{\nz}[0]{n_{\mathbf{z}}}
\newcommand{\Kw}[1]{\mathbf{K}_{#1}}
\newcommand{\Kwt}[1]{\mathbf{K}_{#1}^\top}

\newcommand{\Snu}[1]{\mathbf{S}_{#1}}

\newcommand{\genE}{\boldsymbol{\epsilon}}
\newcommand{\genMu}{\boldsymbol{\mu}}
\newcommand{\genW}{\boldsymbol{\Lambda}}
\newcommand{\genP}{\boldsymbol{\Sigma}}

\newcommand{\tavgnees}{\tilde{\epsilon}_x}
\newcommand{\savgnees}{\tilde{S_x}}
\newcommand{\tavgnis}{\tilde{\epsilon}_z}
\newcommand{\savgnis}{\tilde{S_z}}

\DeclareMathOperator*{\trace}{\mathrm{tr}}

\begin{document}

\title{Kalman Filter Auto-tuning through Enforcing Chi-Squared Normalized Error Distributions with Bayesian Optimization}

\author{Zhaozhong Chen}
\affil{University of Colorado, Boulder, CO 80309, USA} 

\author{Harel Biggie}
\affil{University of Colorado, Boulder, CO 80309, USA} 

\author{Nisar Ahmed}
\affil{University of Colorado, Boulder, CO 80309, USA} 

\author{Simon Julier}
\affil{University of College London, UK}

\author{Christoffer Heckman}
\affil{University of Colorado, Boulder, CO 80309, USA} 






\maketitle

\begin{abstract}
The nonlinear and stochastic relationship between noise covariance parameter values and state estimator performance makes optimal filter tuning a very challenging problem. Popular optimization-based tuning approaches can easily get trapped in local minima, leading to poor noise parameter identification and suboptimal state estimation. Recently, black box techniques based on Bayesian optimization with Gaussian processes (GPBO) have been shown to overcome many of these issues, using normalized estimation error squared (NEES) and normalized innovation error (NIS) statistics to derive cost functions for Kalman filter auto-tuning.While reliable noise parameter estimates are obtained in many cases, GPBO solutions obtained with these conventional cost functions do not always converge to optimal filter noise parameters and lack robustness to parameter ambiguities in time-discretized system models. This paper addresses these issues by making two main contributions.
First, we show that NIS and NEES errors are only chi-squared distributed for tuned estimators. As a result, chi-square tests are not sufficient to ensure that an estimator has been correctly tuned. We use this to extend the familiar consistency tests for NIS and NEES to penalize if the distribution is not chi-squared distributed. Second, this cost measure is applied within a Student-t processes Bayesian Optimization (TPBO) to achieve robust estimator performance for time discretized state space models. The robustness, accuracy, and reliability of our approach are illustrated on classical state estimation problems.
\end{abstract}


\begin{figure}[ht!]
\centering
\includegraphics[width=\linewidth]{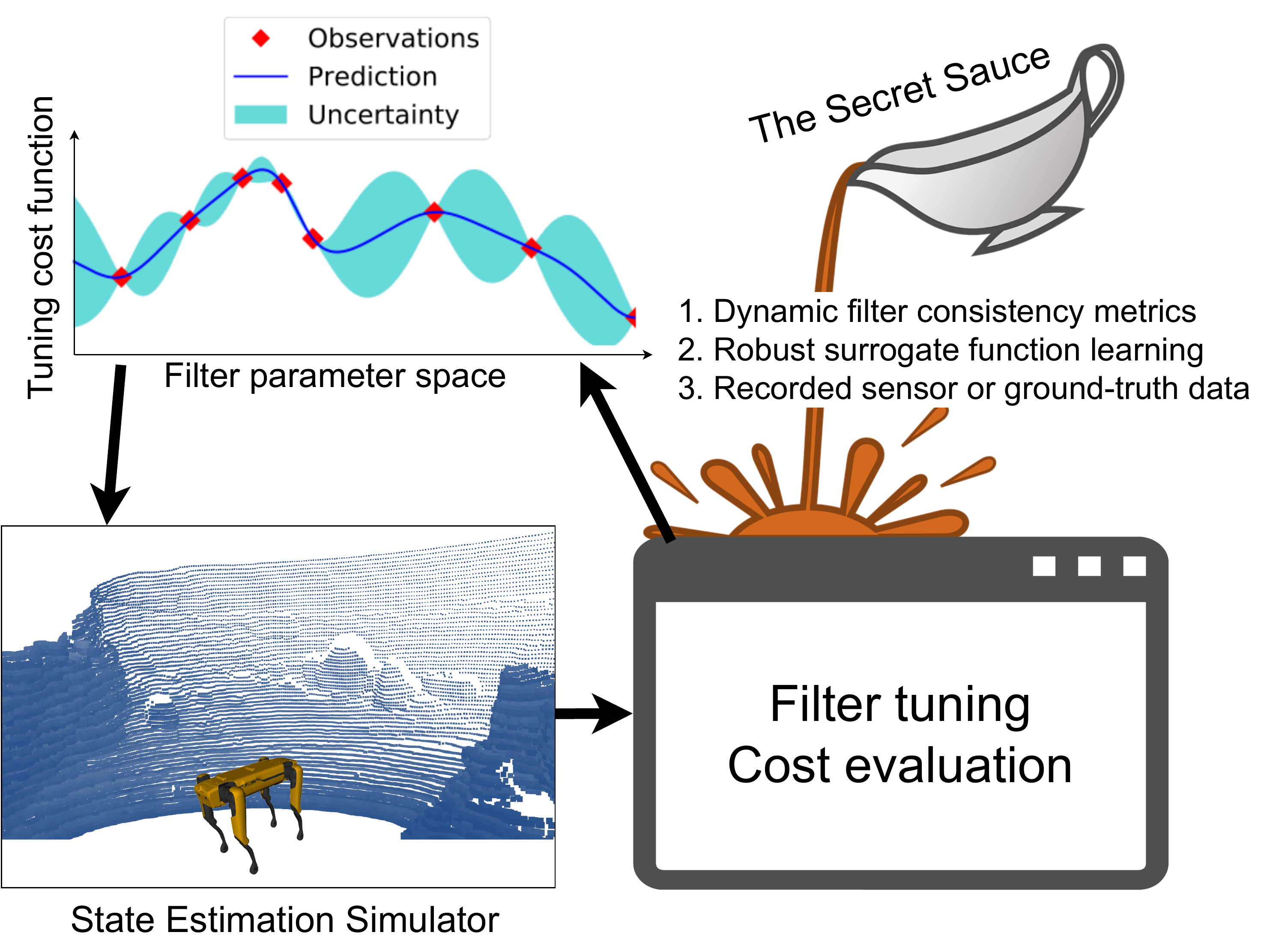}
\caption{We utilize nonparametric Bayesian optimization to optimize the Kalman filter process and measurement noise. Our innovative procedure combines new cost metrics based on Chi-squared tests with T-Process-based Bayesian Optimization for a better evaluation of the filter consistency.}
\label{fig:example1}
\end{figure}

\section{Introduction}

State estimation algorithms are a key component of many autonomous navigation, tracking, and data fusion systems. However, the performance of these estimators is heavily influenced by the choice of the noise parameters within them. 
Most estimation algorithms have two main steps: state prediction followed by measurement update. The state prediction step uses a process model to predict how the state will evolve over time. The measurement update step uses an observation model to incorporate measured quantities into the estimate of the state. Almost all models are wrong, and the errors in these models are treated as random noise. Since most designs assume these errors are drawn from a white zero mean and uncorrelated distribution, only the noise covariances for the prediction and observation models need to be chosen or tuned.


In this paper, we consider the problem of developing an algorithm to automatically tune the noise parameters. We focus specifically on the problem of tuning a Kalman filter, although our work should generalize directly to many other types of estimators. 
Our work makes two main contributions. First, we develop a new measure of statistical consistency. We show that the widely-used NEES/NIS statistics are \emph{only\/} $\chi^2$ distributed random variables when the filter is correctly tuned. When the filter is not tuned (e.g., during the tuning process), the NEES/NIS statistics are sampled from a \emph{generalized\/} $\chi^2$ distribution. This means that optimization metrics based on $\chi^2$ distributions can fail to converge onto correct noise covariance parameters. We use this to derive a new consistency measure and objective functions for tuning algorithms that are robust to this situation. Second, we develop a new tuning algorithm based on Student-t process Bayesian Optimization (TPBO), which overcomes the limitations of previously developed Gaussian process Bayesian Optimization (GPBO) methods by providing additional robustness to noisy objective function evaluations in the noise covariance parameter search.  We illustrate the resulting algorithm in several systems with different dimensionalities. 

The structure of this paper is as follows. 
Section~\ref{sct:preliminaries} presents the problem formulation and preliminaries. Secontion \ref{sct:measures_of_consistency} derives and describes the conventional filter consistency measuring approach. Section~\ref{sct: selection} reviews filter consistency measures typically used for optimization-based black box filter auto-tuning. Section~\ref{sct:new_consistency_measure} describes our new consistency measures for determining whether a filter is tuned correctly, which incorporate information about both the means and variances of the underlying state and measurement error distributions. Section~\ref{sct:bayesopt} describes our new Student's-t process-based Bayesian Optimization auto-tuning algorithm, which provides additional robustness in the noise covariance parameter search process. Section~\ref{sct:experiments} presents comparative results using our new approach and other state of the art algorithms on three different linear filter auto-tuning problems of increasing complexity. Section~\ref{sct:conclusion} concludes the paper.

\section{Prelminaries}
\label{sct:preliminaries}

\subsection{System Overview}

Our approach depends upon adjusting the prediction interval in the Kalman filter. Therefore, it is important to understand the relationship between discrete and continuous time systems. The state of the system at time $t$ is $\x{t}$. The system is described by a continuous-time process model and observation models,
\begin{equation}
\label{eq:con_sys_model}
\begin{aligned}
\dot{\mathbf{x}}_t &= \mathbf{A}_t \x{t} + \mathbf{G}_t \uVec{t} + \boldsymbol{\Gamma}_t \vVec{t}, \\
\z{t} &= \mathbf{H}_t \x{t} + \wVec{t},
\end{aligned}
\end{equation}
where $\uVec{t}$ is the control input, the process noise is the additive white process $\vVec{t}$ with intensity $\V$, and the measurement noise is an additive white noise process $\wVec{t}$ with continuous time intensity $\W$. 

Using time discretization techniques such as Van Loan's method~\cite{mohindergrewalVanLoanMethod2015}, the discrete-time process model describes the evolution of the system from timestep $\kLst$ to $\kCur$:
\begin{equation}
\x{\kCur}=\F{\kCur}\x{\kLst}+\mathbf{B}_{\kCur}\uVec{\kCur}+\vVec{\kCur},
\label{eq:dynModel}
\end{equation}
\noindent where $\uVec{\kCur}$ is the control input (e.g. under zero-order hold) and $\vVec{\kCur}$ is the process noise, which is assumed to be independent with a zero
mean and covariance $\Q{\kCur}$. The observation model is
\begin{equation}
\z{\kCur}=\HM{\kCur}\x{\kCur}+\wVec{\kCur}, \label{eq:measModel}
\end{equation}
\noindent where $\wVec{\kCur}$ is the observation noise. This is assumed to be zero-mean and independent with covariance $\R{\kCur}$.

In this paper, we explore the case that a Kalman filter is used to find the optimal estimate \cite{Kalman-JBE-1961}. Consider the state at time $k$ conditioned on all measurements up to time $j$. The estimate from the Kalman filter is the mean $\xCond{k}{j}$ and the associated covariance $\covCond{k}{j}$. The filter follows the two-stage process of prediction followed by an update. The prediction is:
\begin{align*}
\xCond{\kCur}{\kLst}&=\F{\kCur}\xCond{\kLst}{\kLst} +\mathbf{B}_{\kCur}\uVec{\kCur} \label{eq:kalman1}\\
\covCond{\kCur}{\kLst}&=\F{\kCur}\covCond{\kLst}{\kLst}\Ft{\kCur}+\Q{\kCur}
\end{align*}
\noindent The update is: 
\begin{align*}
\xCond{\kCur}{\kCur}&=\xCond{\kCur}{\kLst}+\Kw{\kCur}\ez{\kCur},\\
\covCond{\kCur}{\kCur}&=\covCond{\kCur}{\kLst}-\Kw{\kCur}\Snu{\kCur|\kLst}\Kwt{\kCur},\\
\Snu{\kCur|\kLst}&=\HM{\kCur}\covCond{\kCur}{\kLst}\HMt{\kCur}+\R{\kCur}\\
\Kw{\kCur}&=\covCond{\kCur}{\kLst}\HMt{\kCur}\Snu{\kCur|\kLst}^{-1} 
\end{align*}
where $\ez{\kCur}=\z{\kCur} - \zCond{\kCur}{\kLst}$ is the \emph{innovation vector\/}. Many other recursive estimation algorithms, including Gaussian mixture models, particle filters, and grid-based estimators have a similar two-step structure.

\subsection{Statistical Consistency}
\label{sbs:consistency}

When the model implemented in the filter precisely matches the system equations, the should filter behave in a statistically consistent manner. Specifically, the filter should obey the following conditions for dynamical consistency~\cite{Bar-Shalom2001}:
\begin{enumerate}
\item The state estimation errors are unbiased,
\begin{equation}
\E{\ex{k}} = \mathbf{0}, \ \forall k
\label{eqn:cons_mean}
\end{equation}
\item The estimator is efficient, 
\begin{equation}
\E{\ex{k}\ex{k}^T} = \covCond{k}{k}, \ \forall k
\label{eqn:cons_cov}
\end{equation}
\item The innovations form a white Gaussian sequence, such that for all times $k$ and $j$,
\begin{align}
\ez{k} &\sim {\cal N}(\mathbf{0},\innovCov{k}{k-1})\notag\\
\E{\ez{k}} &= \mathbf{0}, \notag\\
\E{\ez{k}\ez{j}^T} &= \delta_{jk}\cdot \innovCov{k}{k-1}.
\label{eq:innovation_cov}
\end{align}
\end{enumerate}

\section{Measures of Consistency}
\label{sct:measures_of_consistency}

The goal of tuning is to choose values $\Q{\kCur}$ and $\R{\kCur}$ that meet the consistency conditions. Most tuning approaches use an optimization approach where measures of consistency are introduced, and the noise matrices are adjusted to maximize a measurement of the system performance.

\subsection{Filter Error Statistics}

Two widely used performance measurements  are the \emph{normalized estimation error squared
(NEES)} and the \emph{normalized innovation error squared (NIS)}. The NEES is
computed from
\begin{equation}
\nees{k} = \ex{k}^T \covCond{k}{k}^{-1} \ex{k}, \label{eq:neesDef}
\end{equation}
where $\ex{k} = \x{k} - \xCond{k}{k}$. Because this requires knowledge of the true system state ($\x{k}$), it can only be computed in simulation or in situations where an external validation system is being used. NIS, on the other hand, only depends on the observation sequence and does not require ground truth knowledge. It is computed from
\begin{equation}
\nis{k} = \ez{k}^T \innovCov{k}{k-1}^{-1} \ez{k}, \label{eq:nisDef}
\end{equation}
where $\ez{k}$ is the innovation vector.

If the filter is tuned, it can be shown that the NEES and NIS are $\chi^2$-distributed random variables \cite{Bar-Shalom2001}. We illustrate this for the NEES.  Define the vector-valued quantity
\begin{equation}
    \mathbf{b}_k = \covCond{k}{k} ^{-1/2} \mathbf{\ex{k}}
\end{equation}
Substituting, \eqref{eq:neesDef} can be rewritten as
\begin{equation}
\begin{split}
    \nees{k} &=  \left(\mathbf{b}_k^T\covCond{k}{k} ^{1/2}\right)
    \covCond{k}{k} ^{-1}
    \left(\covCond{k}{k} ^{1/2}\mathbf{b}_k\right)\\
    &=\mathbf{b}_k^T\mathbf{b}_k\\
    &=\sum_{i=1}^{\nx}{b_k(i)^2},
    \end{split}
    \label{eqn:nees_as_x_squared}
\end{equation}
where $b_k(i)$ is the $i$th component of $\mathbf{b}_k$.
Since the filter is tuned, the consistency conditions in \eqref{eqn:cons_mean} and \eqref{eqn:cons_cov} mean that $\nees{k}$ is zero-mean with covariance $\covCond{k}{k}$. Therefore, $\mathbf{b}_k$ is a zero mean, $\nx$-dimensional Gaussian-distributed random variable with covariance equal to the identity matrix. Furthermore, this means that each component $b_k(i)$ is also a zero-mean Gaussian random variable with unit covariance. Therefore, $\nees{k}$ is a $\chi^2$-distributed random variables with $\nx$ degrees of freedom. Similar logic can be used to show that the NIS is $\chi^2$-distributed random variable with $\nz$ degrees of freedom.

Given these insights, the mean values of NEES and NIS are used to evaluate consistency. Their values are given by~\cite{Bar-Shalom2001}
\begin{equation}
\begin{aligned}
    \mathbb{E}[{\nees{k}}]{\approx}\nx,\ \ \
    \mathbb{E}[{\nis{k}}]{\approx}\nz.
\end{aligned}
\label{eqn:nees_nis}
\end{equation}
%
%
%
\subsection{Empirically Measuring Consistency}
\label{sbs:emprical_measures}

In practice, NEES/NIS $\chi^2$ tests are conducted using multiple offline Monte Carlo ``truth model'' simulations to obtain ground truth $\x{k}$ values. For a hypothesized set of filter parameters, $N$ Monte Carlo runs are executed: the truth model is run, observations are collected, the filter predicts and updates and the empirical average of the NEES and NIS are computed. This averaging is often done on a time-step-by-time step basis across all Monte Carlo runs,
\begin{align}
\avgnees{k} &= \frac{1}{N} \sum_{i=1}^{N}{\nees{k}^{[i]}} \label{eq:neesAvgk} \\
\avgnis{k} &= \frac{1}{N} \sum_{i=1}^{N}{\nis{k}^{[i]}}. \label{eq:nisAvgk}
\end{align}
\noindent where the superscript $[i]$ shows this is the quantity from the $i$th run. The estimated values are stochastic because each run is perturbed by random noise. Therefore, given some desired Type I error rate $\alpha$, the NEES, and NIS $\chi^2$ tests provide lower and upper tail bounds
$[l_{\mathbf{x}}(\alpha,N),u_{\mathbf{x}}(\alpha,N)]$ and
$[l_{\mathbf{z}}(\alpha,N),u_{\mathbf{z}}(\alpha,N)]$, such that the Kalman filter tuning is declared to be consistent if, with probability $100(1-\alpha)$
at each time $k$,
\begin{equation}
\label{eq:consistency_ul}
\begin{aligned}
&\avgnees{k} \in [l_{\mathbf{x}}(\alpha,N),u_{\mathbf{x}}(\alpha,N)],  \\
&\avgnis{k} \in [l_{\mathbf{z}}(\alpha,N),u_{\mathbf{z}}(\alpha,N)].
\end{aligned}
\end{equation}

If these conditions are not met, the filter is declared to be inconsistent.  Specifically, if $\avgnees{k}<l_{\mathbf{x}}(\alpha,N)$ or
$\avgnis{k}<l_{\mathbf{z}}(\alpha,N)$, then the filter tuning is ``pessimistic'' (underconfident), since the filter-estimated state error/innovation covariances are too large relative to the true values.  On the other hand, if $\avgnees{k}>u_{\mathbf{x}}(\alpha,N)$ or
$\avgnis{k}>u_{\mathbf{z}}(\alpha,N)$, then the filter tuning is ``optimistic'' (overconfident), since the filter-estimated state error/innovation covariance is too small relative to the true values. Note that as the number of Monte Carlo runs $N$ increases, the stochastic variation in $\avgnees{k}$ and $\avgnis{k}$ decreases, and the bounds become tighter.

We can use the above metric to evaluate the consistency of a filter. However, because this metric is guaranteed to be non-negative, it is not symmetric -- the minimum is zero (if the covariance goes to infinity), and the maximum is unbounded from above (if the covariance matrix becomes singular). One way to overcome this is to use the absolute value of the log of the normalized errors~\cite{chen2018weak,8962573,powell2002automated,morales2008vehicle},
\begin{align}
\label{eq:jnees}
\Jnees&=\left|  \log \left(\frac{\tavgnees}{\nx} \right)  \right|,\\
\label{eq:jnis}
    \Jnis&=\left|  \log \left(\frac{\tavgnis}{\nz} \right)  \right|,
\end{align}
where
\begin{equation*}
    \tavgnees = \frac{1}{T}\sum_{k=1}^{T}{\avgnees{k}},~ \ 
    \tavgnis = \frac{1}{T}\sum_{k=1}^{T}{\avgnis{k}}.
\end{equation*}

We next review existing work on selecting $\Q{\kCur}$ and $\R{\kCur}$.



\section{Process and Observation Noise Selection}
\label{sct: selection}

\subsection{Two-Stage Approach}

There are various existing ways to tune the Kalman filter. Perhaps the simplest approaches use a two-stage divide-and-conquer strategy. The observation noise is first determined using lab-based sensor measurements. The observation noise is held fixed and the process noise is adjusted until the conditions in \eqref{eq:consistency_ul} are satisfied.

However, there are several problems with this two-stage approach. First, with laboratory testing, it is not always possible to model the reaction of sensors in real-world operational environments. For instance, changes in temperature can cause the biases in inertial measurement units (IMUs) to change \cite{altinoz2018determining} which can result in the poor characterization of the observation noise. Additionally, classifying the interactions between noise levels and filter performance is not straightforward, even in the case of linear observation models. It has been theoretically shown that even when the process and observation models are linear, the presence of modeling errors leads to state-dependent noise models which are correlated over time \cite{dette2016optimal, wanninger1998real}. 


\subsection{Autotuning Algorithms}

A more practical alternative is to use auto-tuning algorithms which include: maximum likelihood and Bayesian inference \cite{Bishop2006}, least squares for data processed via Kalman smoothing \cite{barratt2020fitting}, and auto-/cross-correlation analysis \cite{dunik2020covariance}. While all of these methods have their advantages, no single best technique exists for practical use \cite{zhangIdentificationNoiseCovariances2020,dunik2017noise}. These methods are theoretically advantageous for well-defined linear systems where noise models have known structure, and are useful in online settings. Yet, they can also suffer from numerical stability and implementation issues, making them harder to use. Moreover, they are difficult to generalize for non-linear filters, e.g. since the optimal set of noise parameters in linearization-based filters can vary significantly with system state and time \cite{ko2009gp}. 

In this work, we focus on posing the tuning as an optimization problem. We try to identify parameters directly using data from actual rollouts of the system and corresponding filters. This effectively eschews the two-step process; instead, we minimize a customized cost attached to some property of deleterious estimates. Researchers have tried various optimizers such as gradient descent \cite{xia1994adaptive, 9147485}, generalized least squares  \cite{aakesson2007tool}, downhill simplex \cite{powell2002automated}, and EM algorithm \cite{PONCELA2013349}. However, a good initial guess needs to be provided for these methods to prevent them from getting stuck inside of local minima. Such minima are frequent due to inherent nonlinearities in the process and observation models, and because of sampling noise. Therefore, there is considerable interest in exploring more global methods, including genetic algorithms \cite{ting2014tuning,4636432, oshman2000optimal} and Bayesian optimization. The approach in \cite{oshman2000optimal} finds a globally optimal parameter value set is found but is sample-inefficient, requiring thousands of cost function evaluations to determine the optimum.

\subsection{Implicit Time Dependence}
\label{sec:dt_dependence}

We discretize the plant transformation using a sample time of $\Delta t$. As we have previously shown in \cite{chen2021time}, the choice of $\Delta t$ has a significant impact on the performance of the filter. When a single sample time is used, multiple choices of $\Q{\kCur}$ and $\R{\kCur}$ produce consistent-looking results according to $\Jnis$ and $\Jnees$. In \cite{chen2021time}, we demonstrated that the ambiguity could be resolved by evaluating filter performance over several prediction intervals and using the maximum of $\Jnis$ as the cost function (evaluating the worst-case behavior).\footnote{To correctly handle the different prediction intervals, the models were specified in continuous time and converted to discrete-time using Van Loan's method \cite{BrownHwang2012}.} In this work, we leverage all the filter performances and sum the cost from each interval prediction as the cost function to the optimization. 
Specifically, suppose we have $n$ prediction time intervals $\{ \Delta t_1, \cdots , \Delta t_n  \} $. Let $\Jnis(\Delta t_n)$ be the $\Jnis$ value computed when the prediction interval is $\Delta t$. 
 In particular, we used the value
\begin{equation}
    \Jnis= \sum_{i=1}^{n}\Jnis(\Delta t_n)
    \label{eqn:jnis_sum}
\end{equation}

 Theoretically, more prediction intervals can result in better optimization but it can increase the overall computation time. The $n$ value is decided empirically; e.g. as detailed in Sec. \ref{sct:experiments}, for a low-dimension system we use $n = 2$, and for a higher-dimension system we use $n=4$. 
 Although this solution was sufficient to eliminate the ambiguity identified in \cite{chen2021time}, it 
 does not eliminate all ambiguities. In particular, there is a fundamental theoretical issue in the use of NEES and NIS to ensure that a filter has been tuned consistently. We illustrate this in a simple linear example.

\section{New Consistency Measures}
\label{sct:new_consistency_measure}

In this section, we introduce a new consistency measure. We first present an example that illustrates the limitations of $\Jnees$ and $\Jnis$. Then we analyse the underlying reason for these limitations and present a new and more robust measure.

\subsection{Illustrative Example}
\label{sbs:example_consistency}

Consider a linear particle tracking problem. The target state is $\x{} = [x, y, \dot{x}, \dot{y}]^T$, where the motion is corrupted by random accelerations. The particle's position is measured, and the measurements are perturbed by zero-mean, Gaussian-distributed noise. 
\Ignore{
Therefore, the continuous-time process model is
\begin{equation}
\begin{aligned}
&\mathbf{A} =
\begin{bmatrix}
0 & 0 & 1 & 0 \\
0 & 0 & 0 & 1 \\
0 & 0 & 0 & 0 \\
0 & 0 & 0 & 0 \\
\end{bmatrix} \ \ 
\mathbf{G} =
\begin{bmatrix}
0 \\
0 \\
1 \\
1  
\end{bmatrix} \ \
\HM{} =
\begin{bmatrix}
1 & 0\\
0 & 1\\
0 & 0\\
0 & 0\\
\end{bmatrix}^T \\ 
&\mathbf{\Gamma} = 
\begin{bmatrix}
0 & 0\\ 
0 & 0\\
1 & 0\\
0 & 1
\end{bmatrix} \ \
\mathbf{V} =
\begin{bmatrix}
v_0 & 0\\
0 & v_1
\end{bmatrix} \ \
\mathbf{W} =
\begin{bmatrix}
w_0 & 0\\
0 & w_1
\end{bmatrix}.
\end{aligned}
\end{equation}
Using Van Loan's method, the discrete-time models are}
The discrete-time system equations are:
\begin{equation}
\begin{aligned}
    &\mathbf{F} = 
    \begin{bmatrix}
    1 & 0 & \Delta t & 0\\
    0 & 1 & 0 & \Delta t\\
    0 & 0 & 1 & 0\\
    0 & 0 & 0 & 1\\
    \end{bmatrix}
    \ \
    \mathbf{B} = 
    \begin{bmatrix}
    0.5\Delta t^2 \\ 0.5\Delta t^2 \\ \Delta t \\ \Delta t
    \end{bmatrix}
    \\
    &\mathbf{Q} = 
    \begin{bmatrix}
     \frac{\Delta t^3}{3}v_0 & 0 & \frac{\Delta t^2}{2}v_0 & 0\\
     0 & \frac{\Delta t^3}{3}v_1 & 0 & \frac{\Delta t^2}{2}v_1\\
     \frac{\Delta t^2}{2}v_0 & 0 & \Delta t v_0 & 0\\
     0 & \frac{\Delta t^2}{2}v_1 & 0 & \Delta t v_1\\
    \end{bmatrix}
\end{aligned}
\end{equation}
The simulation setup implements this model with noise values $(v_0,v_1,w_0,w_1)=(1, 2, 0.2, 0.1)$ and timestep length $\Delta t= 0.1$.

\begin{figure}[ht!]
\centering\includegraphics[width=80mm]{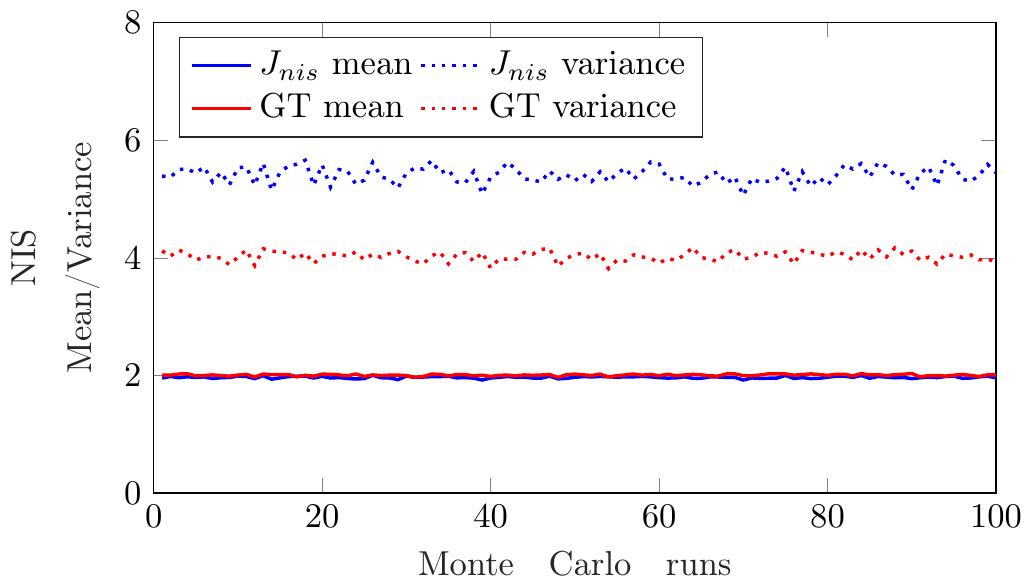}
\caption{Linear tracking example, where ground truth (GT, red) and tuned NIS ($\Jnis$,blue) statistics are shown. We observe that the means are the same but the variances are different which implies the mean is not sufficient for tuning the filter.
}
\label{fig:nis_nees_mismatch}
\end{figure}

Next, consider the case where a filter is implemented using the ground truth values. The mean value of the NIS, computed from \eqref{eq:nisAvgk}, should be $\nz=2$. This is confirmed in Figure \ref{fig:nis_nees_mismatch}, which plots (in red) the average NIS over 100 Monte Carlo runs. However, the figure also shows (in blue) the average NIS statistics for a filter tuned with the values $(v_0,v_1,w_0,w_1)=(0.855,3.000,0.122,0.294)$. Even though these values are very different from the ground truth parameters, the mean value of the tuned NIS is, once again, 2.
%
%
%
This clearly shows that the NIS (and by extension the NEES) values are not sufficient, on their own, to show that a filter is correctly tuned.

\subsection{Limitations of $\Jnees$ and $\Jnis$}

$\Jnees$ and $\Jnis$ both assume that \eqref{eqn:nees_as_x_squared} holds true and these values are $\chi^2$-distributed. However, when the filter is not tuned correctly, two effects can arise. First, the state error and innovation might no longer have a zero mean. Second, the filter incorrectly estimates the covariance of the state and the innovation. As a result, \eqref{eqn:nees_as_x_squared} now needs to include mean $\mu_k(i)$ and variance $w_k(i)$ and becomes 
\begin{equation}
    \nees{k} =\sum_{i=1}^{\nx}{w_k(i)\left(b_k(i)+\mu_k(i)\right)^2}.
\label{eqn:generalized_nees}
\end{equation}
\noindent This random variable, as a sum of squares of independent normal variables, is described by a \emph{generalized} $\chi^2$ distribution. 
Therefore, we need a consistency measure that will be able to differentiate between the generalized $\chi^2$ distribution of a mistuned filter and the $\chi^2$ distribution of a tuned filter. Although complicated sophisticated distribution fit tests could be applied, we find that a simple test, based on the moments of quadratic forms, is sufficient.

\subsection{Moments of a Quadratic Form}


Both \eqref{eq:neesDef} and \eqref{eq:nisDef} are equivalent to computing the expected value of the quadratic form
\begin{equation*}
    \bar{\epsilon}=\mathbb{E}[{\genE\genW\genE^T}],
\end{equation*}
where $\genE$ is a random vector with mean $\genMu$ and covariance $\genP$, the $\genW$ is a is a positive definite symmetric matrix. Taking expectations,
\begin{equation}
    \bar{\epsilon}=\trace[\genW\genP]+\genMu^T\genW\genMu.
    \label{eqn:alpha_mean}
\end{equation}

Consider the case we are computing the NEES. When the filter is tuned correctly, the results in Subsection~\ref{sbs:consistency} mean that $\genW=\genP^{-1}$ and $\genMu=\mathbf{0}$. Substituting into \eqref{eqn:alpha_mean},
\begin{equation}
\label{eq: exp_mea}
\begin{aligned}
\bar{\epsilon}&=\trace[\genP^{-1}\genP]+\mathbf{0}^T\genW\mathbf{0}\\
    &=\trace[\mathbf{I}_n]\\
    &=\nx.
\end{aligned}
\end{equation}
Similiarly, for the NIS values, $\genW=\innovCov{k}{k-1}^{-1}$, $\genMu=\mathbf{0}$, and the expected value is $\nz$.

The quadratic form also lets us compute the covariance. If $\genE$ is Gaussian-distributed, then it can be shown that
\begin{equation}
    \mathbb{E}[{(\epsilon-\bar{\epsilon})^2}]=2\trace[\genW\genP\genW\genP]+
    4\genMu^T\genW\genP\genW\genMu.
\end{equation}
Therefore, when the filter is tuned correctly, the covariance in the NEES is
\begin{equation}
\begin{aligned}
    \mathbb{E}[{(\epsilon-\bar{\epsilon})^2}]&=2\trace[\genP^{-1}\genP\genP^{-1}\genP]+
    4\mathbf{0}^T\genW\genP\genW\mathbf{0}\\
    &=2\trace[\mathbf{I}_n^2]\\
    &=2\nx.
    \label{eqn:alpha_cov}
\end{aligned}
\end{equation}
and the covariance in the NIS is $2\nz$.

We can see the relevance of these values in Fig.~\ref{fig:nis_nees_mismatch}, which also plots the covariance values of the NIS computed over the Monte Carlo runs. For the correctly tuned filter, the covariance is 4, which is $2\nz$. However, for the mistuned filter, the covariance is about 5.75. This clearly shows that the tuned NIS value follows a standard $\chi^2$ distribution.

Using these insights, we introduce new consistency measures which ensure that both the mean and covariance values are correct.

\subsection{New Consistency Measure}

Our new NEES measure has the form
\begin{equation}
    \begin{aligned}
    & \Cnees =  \left| \log \left(\frac{\tavgnees}{\nx} \right)  \right| + \left| \log \left(\frac{\savgnees}{2\nx} \right)  \right|,
    \\
    & \savgnees = \frac{1}{T(N-1)} \sum_{k=1}^{T} \sum_{i=1}^{N}({\nees{k}^i}- \avgnees{k}^i)^2.
    \end{aligned}
\end{equation}
\\
\indent Similarly, our new NIS measure has the form.
\begin{equation}
    \begin{aligned}
    \label{eq:cnis}
    & \Cnis =  \left| \log \left(\frac{\tavgnis}{\nz} \right)  \right| + \left| \log \left(\frac{\savgnis}{2\nz} \right)  \right|,
    \\
    & \savgnis = \frac{1}{T(N-1)} \sum_{k=1}^{T} \sum_{i=1}^{N}\left({\nis{k}^i}- \avgnis{k}^i\right)^2.
    \end{aligned}
\end{equation}
Considering the implicit time dependence property  mentioned in Section \ref{sec:dt_dependence} and eq. \eqref{eqn:jnis_sum}, and given multiple time discretization intervals $\Delta t_n$, our final NEES/NIS metrics have the form
\begin{equation}
\label{eq: sum_cnis}
\begin{aligned}
    & \Cnees = \sum_{i=1}^{n} \Cnees(\Delta t_n), \\
    & \Cnis = \sum_{i=1}^{n} \Cnis(\Delta t_n).
\end{aligned}    
\end{equation}

\section{Bayesian Optimization-based Tuning}
\label{sct:bayesopt}




Now that we have more suitable measures of consistency, the next task is to develop an algorithm that will choose the noise parameters to optimize the desired consistency measure. 
However, this leads to a very challenging optimization problem. The relationship between noise parameters and $\Cnis$ or $\Cnees$ can be highly nonlinear with many local minima. This difficulty is compounded by the fact that these quantities are computed empirically with a finite number of samples. Therefore, each computed value will be stochastic. A principled way to address both multiple local minima and noisy cost function evaluations is to use \emph{Bayesian optimization\/} (BO) \cite{pelikan1999boa}. This poses optimization as a probabilistic search problem in which the goal is to find the global minimum within a bounded region. The system maintains a probabilistic estimate of the cost function within this search region, and the optimizer refines and uses this estimate as it searches for the optimal solution.



\subsection{Student-t Process Bayesian Optimization Theory}
\label{subsct:BA_theory}

Consider the objective function $y:\bovarspace \rightarrow \mathbb{R}$ which maps a point $\bovar$ from the $d$-dimensional space $\bovarspace \in \mathbb{R}^d$ to the real value. 
We assume that the value of each elements of $i$ of $\bovar$ is bounded in a finite region. Specifically, for 
the $i$th component of $\bovar$, $\bovar(i) \in [\bovar(i)_l, \bovar(i)_u]$. The goal is to find $\bovar^* \in \bovarspace$ which minimizes $y$.

\indent There are the two main components in the Bayesian optimization algorithm: (1) the surrogate model $\fs$, which encodes statistical beliefs about $\fy$; and (2) the acquisition function which is used to intelligently guide the search for $\bovar^*$ using $\fs$.

\subsubsection{Surrogate Model}
\indent This is a probabilistic approximation of the objective function $y$ over the search space. In general, $y$  is expensive to evaluate for filter tuning because data from a candidate filter configuration (i.e. with candidate noise covariance values) must be collected either from ground truth data for NEES-based evaluation or from recorded sensor data logs for NIS-based evaluation. On the other hand, a surrogate model of $y$ is easier to evaluate and can provide sufficient information to guide a search toward optimum covariance parameters that are informed by data. Moreover, the surrogate model can embed the locations of multiple local minima without trapping the search process. 

To this end, we use a nonparametric regression model based on a Student's-t process to construct the surrogate function from sampled data runs using a candidate filter tuning.
By definition, a Student's-t process is a stochastic process such that every sample $\mathbf{q}$ from the process has the multivariate Student-t joint distribution
\begin{equation}
    y(\mathbf{q}) \sim \mathcal{TP}(v, \Phi(\mathbf{q}), k(\mathbf{q},\mathbf{q}')).
\end{equation}
The mean function is $\Phi(\mathbf{q})$, the kernel function is $k(\mathbf{q})$ and the parameter $v>2$ controls how heavy-tailed the process is. Smaller values of $v$ correspond to heavier tailed distributions, with higher  probabiltiies of extreme values. On the other hand, as $v \rightarrow \infty$, the process converges towards a Gaussian process (GP) with light tails.

The TP is attractive because it provides some extra benefits over GP surrogate models that are more commonly used, without incurring more computational cost. For example, the predictive covariance for the TP explicitly depends on observed $y$ data values; this is a useful property which the Gaussian process lacks.
Furthermore,  distributions over the cost function $y$ may in general be heavy-tailed, so it is better to use TP to ``safely'' model their behaviors \cite{shah2013bayesian}. Similarly, every finite collection of TP samples $\mathbf{q}_{1:n} = (\mathbf{q}_1, \mathbf{q}_2, \cdots, \mathbf{q}_n)$ has a multivariate Student-t distribution, 
\begin{equation}
    y(\mathbf{q}_{1:n}) \sim MVT_n(v, \Phi(\mathbf{q}_{1:n}), K),
    \label{eq:mvt}
\end{equation}
%
\noindent where $K$ is the covariance matrix consisting of kernel function $k$ \cite{Rasmussen2006,minasny2005matern,genton2001classes} evaluations,
\begin{equation} \label{cov_matrix}
    K = 
    \begin{bmatrix}
    k(\mathbf{q}_1, \mathbf{q}_1) & k(\mathbf{q}_1, \mathbf{q}_2) & \cdots & k(\mathbf{q}_1, \mathbf{q}_n) \\
    k(\mathbf{q}_2, \mathbf{q}_1) & k(\mathbf{q}_2, \mathbf{q}_2) & \cdots & k(\mathbf{q}_2, \mathbf{q}_n) \\
    \vdots & \vdots & \vdots & \vdots \\
    k(\mathbf{q}_n, \mathbf{q}_1) & k(\mathbf{q}_n, \mathbf{q}_2) & \cdots & k(\mathbf{q}_n, \mathbf{q}_n)
    \end{bmatrix}.
\end{equation}

As more and more samples are taken, more and more information becomes available about the objective function. The surrogate model is updated in two ways to exploit this new information. First, the newly sampled $\bovar$ and $y$ values are added to the vector $\mathbf{q}_{1:n}$ and $y(\mathbf{q}_{1:n})$. Second, the hyperparameters of the kernel function $k$ are re-estimated from the data. The updated surrogate model is then used to compute the acquisition function, which is used to select the next sample $\bovar$ for evaluation of $y$. 

\subsubsection{Acquisition Function}

This queries the surrogate model to predict where the next sample should be taken. Many different acquisition functions have been proposed in the Bayesian optimization literature, varying in the kind probabilistic information they exploit for particular applications. In this work we use \emph{expected improvement} (EI), which selects the point which has the highest probability of being the optimal solution. To achieve this, we first define the improvement,
\begin{equation}
    g(\mathbf{q}_{n+1}, \mathbf{q}^*_{1:n}) = \max(0, y_n(\mathbf{q}^*_{1:n}) - y(\mathbf{q}_{n+1})).
\end{equation}
This is non-zero only if $\mathbf{q}_{n+1}<\mathbf{q}^*_{1:n}$.
Since we only have access to the surrogate function, the improvement is stochastic. Therefore, we use the \emph{expected improvement} \begin{equation} \label{expected_improvement}
    \mathrm{EI}_n(\mathbf{q}) = \mathbb{E}_n[g(\mathbf{q}, \mathbf{q}^*_{1:n}) \ | \ \mathbf{q}_{1:n}, \mathbf{y}(\mathbf{q}_{1:n}) ],
\end{equation}
\noindent where $E_n[\cdot]$ is the expectation based on current posterior distribution, given by the current $MVT$ surrogate model. The sample with the highest expected improvement is found from
\begin{equation}
    \mathbf{q}_{n+1} = \argmax\limits_{\mathbf{q}} \mathrm{EI}_n(\mathbf{q}).
    \label{eqn:next_q}
\end{equation}

There are two main advantages to the $\mathrm{EI}$ measure. First it has a clear intuitive meaning, 
Second, for the TP surrogate model, a closed form solution exists for \eqref{expected_improvement}~\cite{shah2013bayesian, tracey2018upgrading}:
\begin{equation} 
\begin{split}
    \mathrm{EI}_n(\mathbf{q}) &= \left(y_n(\mathbf{q}^*_{1:n}) - u\right)\Psi(z) + \\
    &\qquad\frac{v}{v-1}\left(1 + \frac{z^2}{v}\right)\sigma \psi(z).
    \end{split}\label{eqn:EI_function}
\end{equation}
See Appendix~\ref{adx:mvt_ei} for more details.

Given this closed for solution, we still need to solve \eqref{eqn:next_q}. Although this is another optimization problem, it is much easier to solve than the original optimization problem. We use \emph{DIRECT}, which is a derivative free and deterministic nonlinear global optimization algorithm that is widely used for Bayesian optimization via nonparametric surrogate model regression~\cite{finkel2003direct}. Once $\mathbf{q}$ has been determined, the cost objective function is evaluated and the surrogate model is updated. The acquisition function is used to identify the next sample and this process repeats until the termination criteria are met. We use maximum iteration or minimum observation change between two iterations.


\subsection{Tuning Algorithm Summary} \label{subsct: summary}



Algorithm \ref{alg:BayesOpt} summarizes the \BO{} procedure for Kalman filter tuning. In addition to the revised consistency measure, it also uses multiple timesteps and the summation scheme from Eq. \eqref{eq: sum_cnis}. Algorithm \ref{alg:compute_tpbo_entry} details the computation of an individual TBPO sample using a data generator function $\Call{DataGenerator}{}$ (ground truth data for NEES, sensor data for NIS), and calls the metric evaluation function shown in Algorithm \ref{alg:compute_cnis_entry}. 
Note that while either $C_{NEES}$ or $C_{NIS}$ could be used in the \BO{} auto-tuning method, the latter is arguably easier to implement in most real applications. This is because the $C_{NIS}$ only requires single-run or multi-run recorded sensor data logs instead of the multiple ground truth state data logs required by $C_{NEES}$, which are generally expensive to acquire either via high-fidelity truth model simulations or high-fidelity sensors.  
Thus, in Algorithms \ref{alg:compute_tpbo_entry} and \ref{alg:compute_cnis_entry} and for the remainder of the paper, $C_{NIS}$ is used to describe and examine properties of the \BO{} auto-tuning method. 

 \begin{algorithm}[H]
            \caption{\BO{} for Kalman filter tuning}
            \label{alg:BayesOpt}
            \begin{algorithmic}[1] 
                    \State Initialize TP seed data $\left\{\bovar_s, \fy_s\right\}_{s=1}^{N_{seed}}$ and hyperparameters $\Theta$
                    \While {$n < N$}
                        \State $\bovar_j = \argmax_{\bovarspace} \fa$ 
                        \State $y(\bovar_j) \leftarrow \Call{computeTPBOsample}{\bovar_j, \Delta{T}_{1,\cdots,m}}$ 
                        \State Add $y(\bovar_j)$ to $\fv(Q)$, $\bovar_j$ to $Q$, and update $\Theta$ \label{ln:testing}
                    \EndWhile \\
                    \Return $\bovar^* = \arg \min_{\bovar_j \in Q} \fv(\bovar_j)$
            \end{algorithmic}
        \end{algorithm}

\begin{algorithm}
\caption{Compute TPBO sample.}
\label{alg:compute_tpbo_entry}
\begin{algorithmic}
\Procedure{computeTPBOsample}{$\bovar, \Delta{T}_{1,\cdots,m} $}
\State $Z \leftarrow \Call{DataGenerator}{}$
\For{$\Delta{T} \in  \{\Delta{T}_1, \cdots, \Delta{T}_m \}$}
\State $\Cnis^i \leftarrow \Call{ComputeCNIS}{\bovar, \Delta{T}_i, Z}$
\EndFor
\State \Return $ \sum \left(\Cnis^i\right)$
\EndProcedure
\end{algorithmic}
\end{algorithm}

\begin{algorithm}
\caption{Compute $C_{NIS}$ for each sample.}
\label{alg:compute_cnis_entry}
\begin{algorithmic}
\Procedure{ComputeCNIS}{$\bovar, \Delta{T}, Z$}
\State Initialize Kalman Filter $\mathbf{K}(\bovar, Z)$ 
\For{$t \in \Delta{T}$}
\State Iterate $\mathbf{K}$ using timestep $t$
\State $\Cnis \leftarrow$ from Eq.\ $\eqref{eq:cnis}$
\EndFor
\State \Return $\Cnis$
\EndProcedure
\end{algorithmic}
\end{algorithm}

\section{Experiments}
\label{sct:experiments}

In this paper, we present filter auto-tuning results on three systems that illustrate different aspects of our approach:
\begin{itemize}
    \item \textbf{Mass-spring-damper}: Here the state consists of the position and velocity of the mass in the system. The process and measurement noise are both 1D. We use the 1D example to highlight our method's ability to predict the noise parameter precisely. Additionally, we optimize over multiple ground truth pairs to demonstrate that our system is robust to different noise parameters, even in the most challenging situations where these are large.
    \item \textbf{2D tracking system}: In the 1D system, tuned estimates obtained via $C_{NIS}$ and the conventional $J_{NIS}$ cost functions are quite similar, and therefore we cannot see the benefits of the new $C_{NIS}$ cost function. However, the 2D tracking system shows that $C_{NIS}$ provides statistically consistent results in higher dimensional problems, while again allowing for evaluation against a small set of ground truth parameters. 
    \item \textbf{Cascade mass-spring-damper system}: We cascade three mass-spring-dampers into six state dimensions to demonstrate consistency in higher dimensions. This problem allows us to show the necessity of the variance term in the new cost function, along with the increased significance of using multiple time discretization intervals.
\end{itemize}

We compare the results of four auto-tuning strategies in each case. The first one is the proposed \BO{} algorithm with the $\Cnis$ cost function. To assess the improvements of the new cost function with respect to our previous work, we compare it to \BO{} with the $\Jnis$ cost function. 
Next, we evaluate the benefits of the Student's-t process surrogate modeling and multi-time scale sampling approaches in TPBO by comparing to the previously developed GPBO method developed in \cite{chen2018weak}, using the $\Jnis$ cost function and $\Delta t = 0.1$ as the only discretization sample time. Finally, we highlight the benefits of our algorithm by comparing it against the state-of-the-art 
downhill simplex auto-tuning algorithm developed by Powell \cite{powell2002automated}, using $\Cnis$ as the cost function and also taking advantage of different $\Delta t$ discretization sample times to ensure a fair comparison. 
After the optimization converges for each method across 120 Monte Carlo simulation runs, we provide a series of evaluations to compare the filters: the true cost function at the converged solution; filter state estimation error accuracy; filter dynamic consistency, i.e. whether the estimated states satisfy eq. \eqref{eq:consistency_ul}; and BO surrogate model visualizations, to demonstrate the solution search process. 
\begin{figure*} [ht!]
\centering
\subfloat[]{
\includegraphics[width=0.47\textwidth]{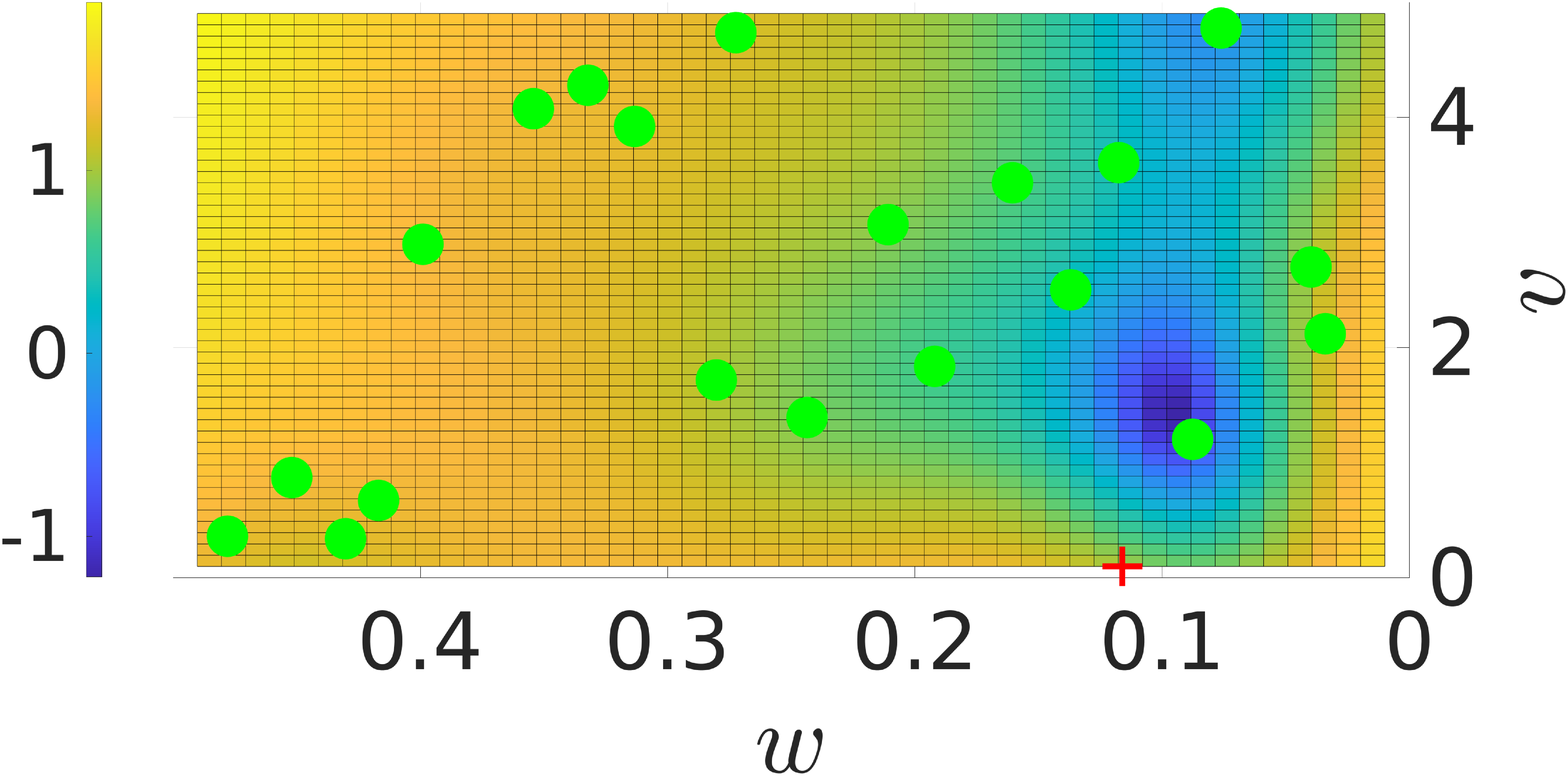}
\label{fig:msp_bayesopt_a}
}
\subfloat[]{
\includegraphics[width=0.47\textwidth]{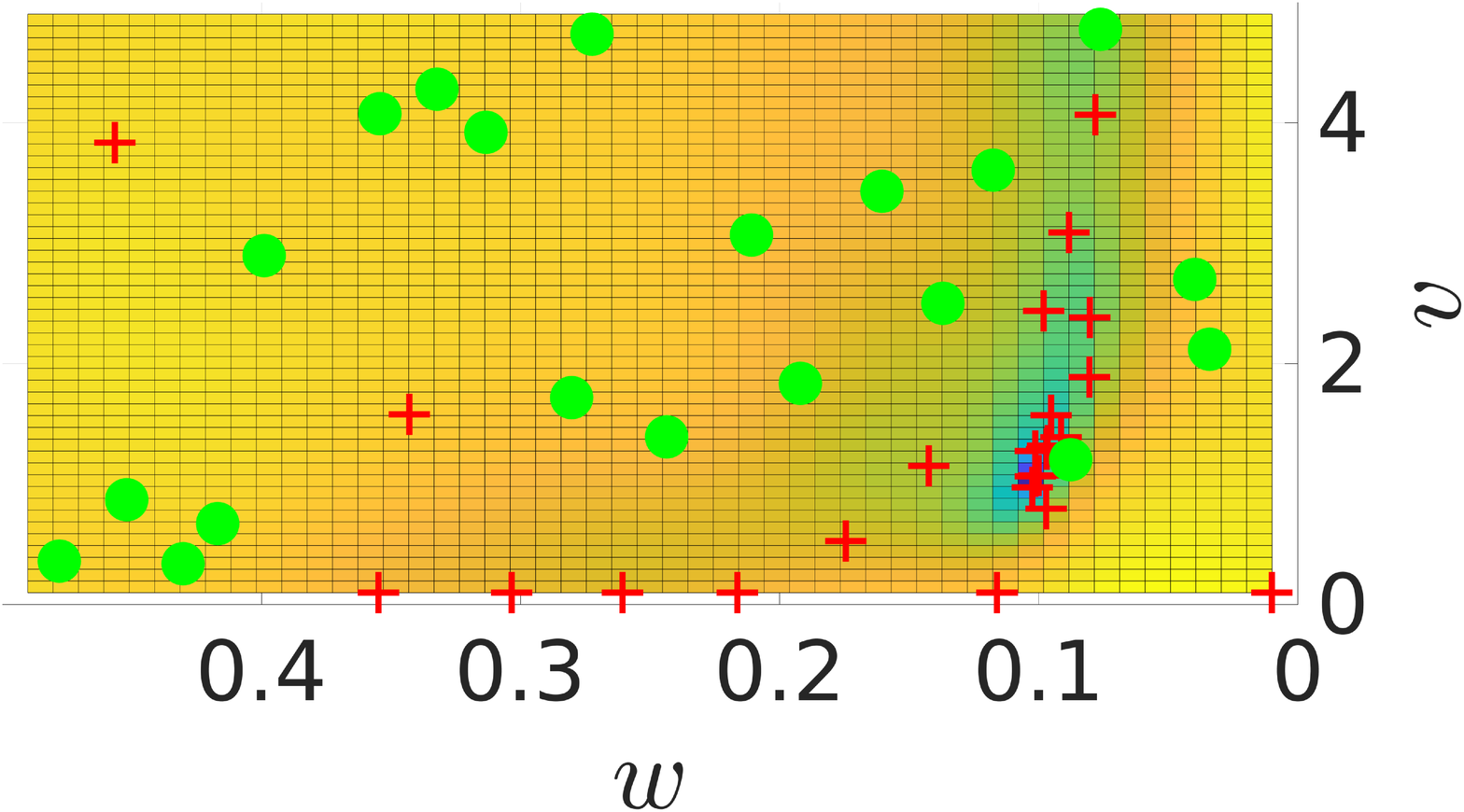}
\label{fig:msp_bayesopt_b}
}\\
\subfloat[]{
\includegraphics[width=0.47\textwidth]{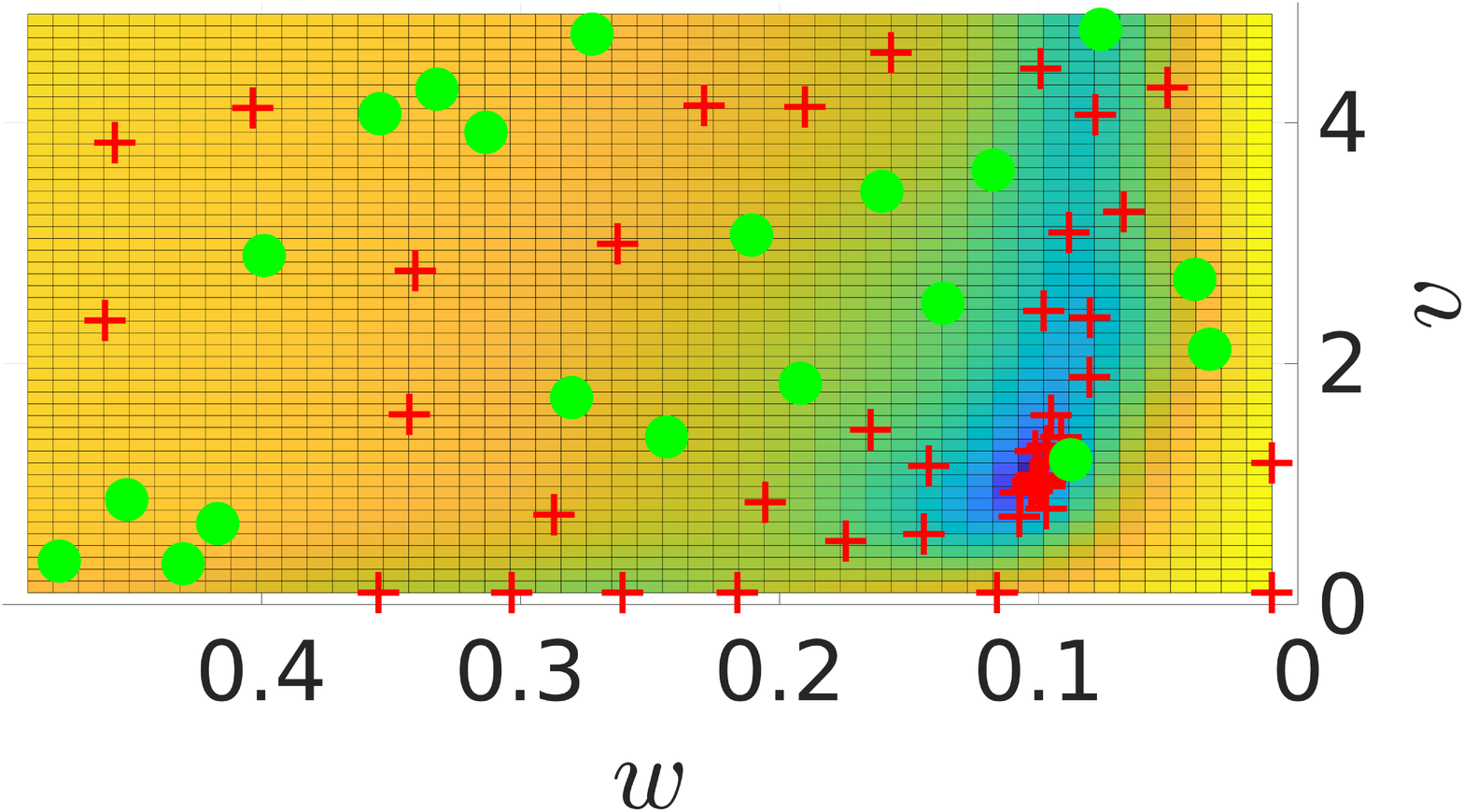}
\label{fig:msp_bayesopt_c}
}
\subfloat[]{
\includegraphics[width=0.47\textwidth]{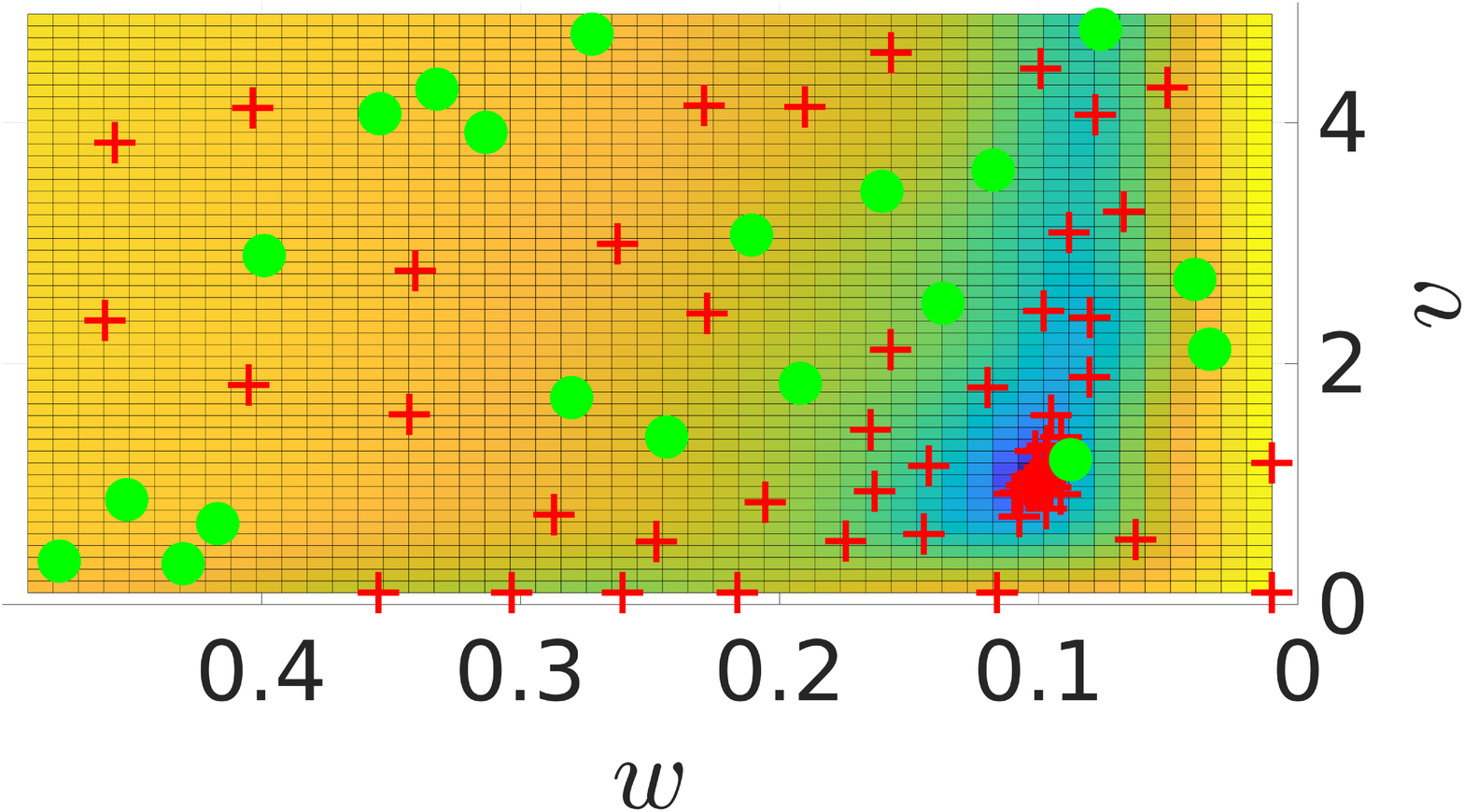}
\label{fig:msp_bayesopt_d}
}
\caption{ \BO{} surrogate model for $C_{NIS}$ cost (color map), showing initial random sample points (green dots) and best estimate (red crosses) inferred by \BO{} as the number of iterations increases from the start (a) to finish (d), where sampled points converge near ground truth $v=1$ and $w=0.1$ after 70 iterations. }
\label{fig:msp_bayesopt}
\end{figure*}

\begin{figure*} [ht!]
\centering
\subfloat[]{
\includegraphics[scale=0.94]{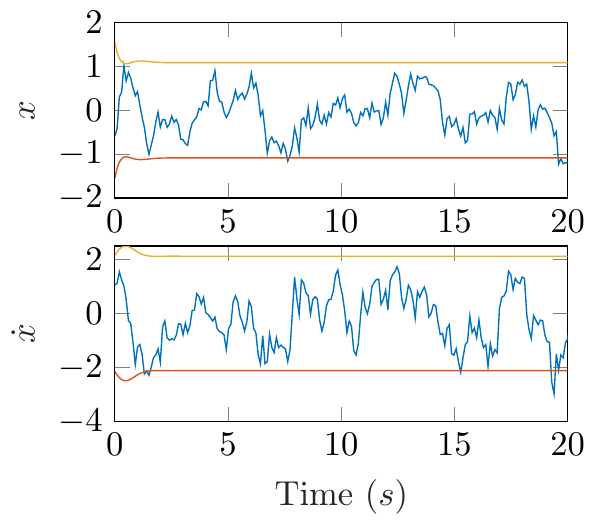}
\label{fig:con_check_msd}
}
\subfloat[]{
\includegraphics[scale=0.94]{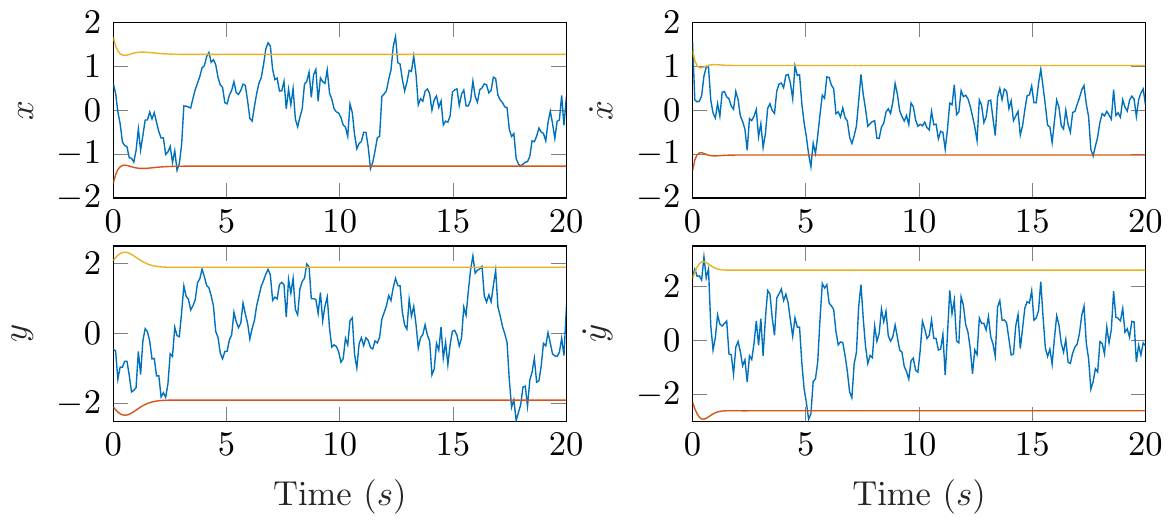}
\label{fig:con_check_tracking2d}
}
\caption{State errors (vertical axis) vs. time (horizontal axis) for mass-spring-damper (a) and 2D tracking (b). Orange lines: $2\sigma$ bounds; blue line: error between estimated and true states in each step of the Kalman Filter. If the system is consistent, around 95$\%$ of state errors should be within the $2\sigma$ range.} 
\label{fig:consistency_check}
\end{figure*}

\subsection{Mass-spring-damping system}
\label{sec:single_msp}

In this example, we introduce the mass-spring-damper model and explore the convergence of our method during the optimization process. First, we run multiple optimizations using the same ground truth and evaluate if the convergence is consistent over multiple runs. Second, we investigate the effectiveness of \BO{} on different sets of ground truth system values.

The classical mass-spring-damping system state is the mass position and velocity $\mathbf{x} = [x,\dot{x}]$. The continuous time state space model is
{\allowdisplaybreaks
\abovedisplayskip = 2pt
\abovedisplayshortskip = 2pt
\belowdisplayskip = 2pt
\belowdisplayshortskip = 2pt
\begin{align*}
\mathbf{A} =
\begin{bmatrix}
0 & 1 \\
-\frac{k}{m} & -\frac{b}{m} 
\end{bmatrix}, \ \ 
\mathbf{G} &=
\begin{bmatrix}
0 \\
\frac{1}{m} 
\end{bmatrix}, \ 
\HM{} =
\begin{bmatrix}
1 & 0  
\end{bmatrix}, \ 
\mathbf{\Gamma} = 
\begin{bmatrix}
0 \\ 
1  
\end{bmatrix}, \\ 
\mathbf{V} &= v \ \ \ , \ \ \
\mathbf{W} = w.
\end{align*}
}
where $m = \SI{1}{\kilo\gram}$ is the mass, $k=\SI{1}{\newton\per\second}$ is the spring constant, and $b=\SI{0.2}{\newton\second\per\metre}$ is the damping constant. We assume the a sinusoidal external force as an input, 1D velocity process noise, and 1D position measurement noise. 
We time discretize this model (see Appendix~\ref{adx:cont_to_disc} for details) and then use \BO{} to optimize the filter noise intensities $v$ and $w$.\\

\begin{table*}[!ht]
\centering
\caption{Mass-spring-damping Auto-tuning Results.}

\begin{tabular}{ *3c | *2c | *2c | *2c | c}
\toprule
& \multicolumn{2}{c}{\BO{} with $\Cnis$} & \multicolumn{2}{c}{\BO{} with  $\Jnis$} & \multicolumn{2}{c}{GPBO from \cite{chen2018weak}} & \multicolumn{2}{c}{Downhill Simplex from \cite{powell2002automated}}&
\multicolumn{1}{c}{Groundtruth}\\
& $v$ & $w$ & $v$ & $w$ & $v$ & $w$ & $v$ & $w$ \\
\midrule
Median & \textbf{1.004} & \textbf{0.0998} & 1.017 & 0.0995  &3.019 &0.146  &0.29 & 0.028 & $v$ = 1 \\
Variance & \textbf{0.003} & 3.13e-6 &0.005 &\textbf{3.11e-6} &2.362 &1.33e-4 &0.39 &0.003 &  $w$ = 0.1\\
Mean & \textbf{1.0189} & \textbf{0.0997} & 1.030 & 0.0993  &3.046 &0.0976  &0.602 & 0.018 \\
\bottomrule
\end{tabular}
\label{table:mass_spring_damping_result}
\end{table*}

\indent The control input for the discretized system is $\mathbf{u}_t = 2 \cos(0.75t)$ (discretized with zero-order hold). 
For the two sampling rates required by \BO{}, two $\Delta t = [0.1,0.5]$ are used with a simulation episode time of $T = 200\Delta t$. $N = 120$ Monte Carlo simulations were used in the Bayes optimization methods. Increasing the value of $N$ can make $\Cnis$ more robust at the expense of computation time. Several other key parameters had to be set for \BO{} and GPBO. For the kernel function, the Mat\'ern Kernel \cite{minasny2005matern}, with automatic relevance determination (ARD), was used. ARD uses independent parameters for every dimension of a given problem. As such, we also treat the optimized parameters as independent within BO. 
The kernel parameter $\nu$ describes how many times the kernel function is differentiable; in our experiments, we use $\nu = 3$ according to the convention in \cite{minasny2005matern}. 
For the remaining parameters such as the kernel mean, kernel hyperparameter re-learn iteration number, and the acquisition function optimization number, we found that the default values from the Bayesian optimization library \cite{JMLR:v15:martinezcantin14a} were sufficient. For the Downhill Simplex method, there are four main parameters: reflection ($re$), expansion ($ex$), contraction ($co$), and full contraction ($fc$). Here, we used $re = 1$, $ex = 1$, $co = 0.5$, $fc = 0.5$. \\

\indent For the fixed ground truth validation, the ground truth is $[v,w] = [1,0.1]$, The BO methods search ranges are $w\in[0.01, 0.5]$ and $v\in[0.1,5]$. We run each optimization 50 independent times. We record the optimized $[v,w]$, and compute the median, variance, and mean of the 50 runs. These data are presented in Table \ref{table:mass_spring_damping_result}. The median and mean of the optimized parameters are expected to be close to the ground truth with only a minor variance. In Figure \ref{fig:msp_bayesopt}, we present the evolution of the \BO{} surrogate model during different iterations of the $C_{NIS}$ optimization process for a single optimization run. It can be seen the proposed method is efficient in predicting and finding the optimized parameters.\\

\begin{figure}[t]
\centering
\includegraphics[width=0.45\textwidth]{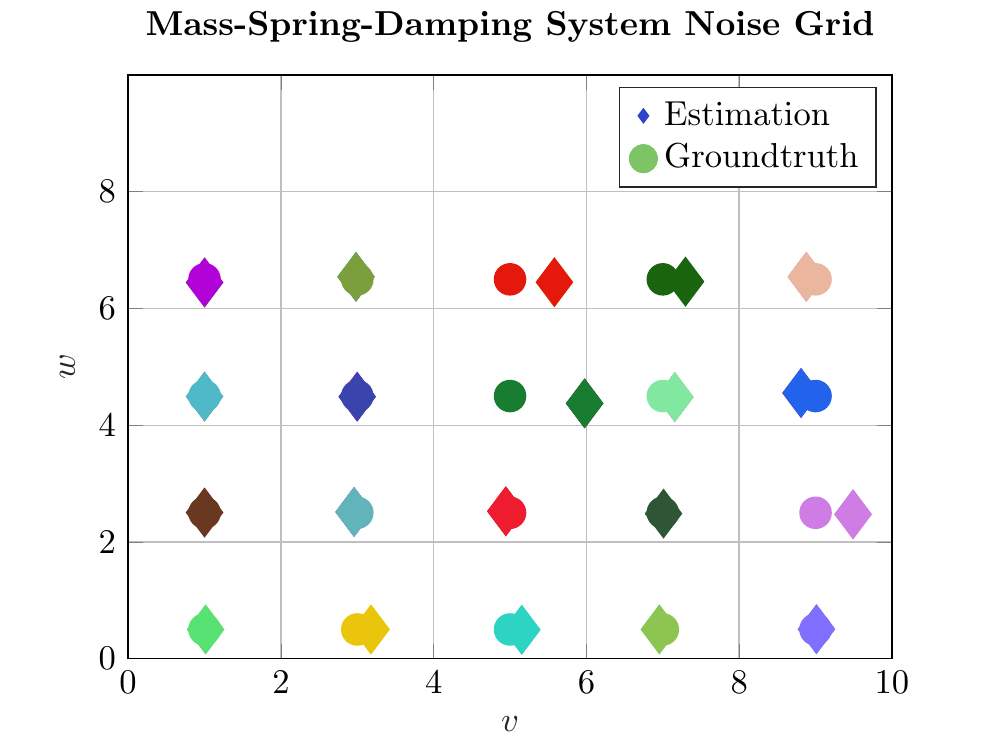}
\caption{Various ground truth pairs ($v$,$w$) for the mass-spring-damping system (circles) and \BO{} tuning results (diamonds). In this case, $w$ is highly observable which results in more system sensitivity than $v$. The results are almost all close to ground truth; even in tough cases such as $v = 5, w = 4.5$, the \BO{} result still passes the $\chi^2$ test.}
\label{fig:multiple_gt}
\end{figure}

\indent For extended validation, multiple ground truth values were also examined on a 5$\times$4 grid of noise parameters defined as follows: $v = [1,3,5,7,9]$ and $w = [0.5,2.5,4.5,6.5]$. 
We increase the BO search range of $w$ and $v$ to $w=[0.1,10]$ and $v=[1,20]$ to match the dynamics of the problem. The initial sample is set to 40 and the iteration count is set to 160. 
For each ground truth pair on the 5$\times$4 grid, we run the optimization once and record the final optimization result. 
Figure \ref{fig:multiple_gt} displays the ground truth value grid (circles) and each corresponding optimization result (diamond). From this we can see in most cases the estimated parameters are close to the ground truth values. Even in test cases where the noise is significantly large (e.g. $v=9,w=6.5$) the BO result is still robust. However, we find that when the true noise parameters are large, it is better to increase the TPBO initial sample number and iterations to ensure good results. 


\subsection{2D Tracking}
\label{sct:2D tracking}

In this section, we show that \BO{} is robust in higher dimensional problems for a system with 4 noise parameters. We implement a 2D tracking system where the target state is $\x{} = [x, y, \dot{x}, \dot{y}]^T$. We assume the same control input for $\dot{x}$ and $\dot{y}$ as in the previous mass-spring-damper system, and add white Gaussian process noise to these states. Additionally, we add white Gaussian measurement noise to a position sensor on $x$ and $y$. Thus, our continuous system model is defined by Eq. \eqref{eq: def_2d_tracking_con} and the closed form discrete time model is Eq. \eqref{eq: def_2d_tracking_dis}. In the continuous time system model, we define the ground truth process and measurement noise parameter $v_0 = 1, v_1 = 2$ and $w_0= 0.2, w_1=0.1$.\\
{\allowdisplaybreaks
\abovedisplayskip = 2pt
\abovedisplayshortskip = 2pt
\belowdisplayskip = 2pt
\belowdisplayshortskip = 2pt
\begin{align}
\label{eq: def_2d_tracking_con}
&\mathbf{A} =
\begin{bmatrix}
0 & 0 & 1 & 0 \\
0 & 0 & 0 & 1 \\
0 & 0 & 0 & 0 \\
0 & 0 & 0 & 0 \\
\end{bmatrix}, \ 
\mathbf{G} =
\begin{bmatrix}
0 \\
0 \\
1 \\
1  
\end{bmatrix}, \ 
\HM{} =
\begin{bmatrix}
1 & 0\\
0 & 1\\
0 & 0\\
0 & 0\\
\end{bmatrix}^T, \\ 
&\mathbf{\Gamma} = 
\begin{bmatrix}
0 & 0\\ 
0 & 0\\
1 & 0\\
0 & 1
\end{bmatrix}, \ 
\mathbf{V} =
\begin{bmatrix}
v_0 & 0\\
0 & v_1
\end{bmatrix}, \
\mathbf{W} =
\begin{bmatrix}
w_0 & 0\\
0 & w_1
\end{bmatrix}
\end{align}
\begin{align}
\label{eq: def_2d_tracking_dis}
    &\mathbf{F} = 
    \begin{bmatrix}
    1 & 0 & \Delta t & 0\\
    0 & 1 & 0 & \Delta t\\
    0 & 0 & 1 & 0\\
    0 & 0 & 0 & 1\\
    \end{bmatrix},
    \ \
    \mathbf{B} = 
    \begin{bmatrix}
    0.5\Delta t^2 \\ 0.5\Delta t^2 \\ \Delta t \\ \Delta t
    \end{bmatrix},
    \\
    &\mathbf{Q} = 
    \begin{bmatrix}
     \frac{\Delta t^3}{3}v_0 & 0 & \frac{\Delta t^2}{2}v_0 & 0\\
     0 & \frac{\Delta t^3}{3}v_1 & 0 & \frac{\Delta t^2}{2}v_1\\
     \frac{\Delta t^2}{2}v_0 & 0 & v_0\Delta t  & 0\\
     0 & \frac{\Delta t^2}{2}v_1 & 0 & v_1\Delta t\\
    \end{bmatrix}.
\end{align}
}
\noindent We apply the four optimization methods for 50 independent trials and the results are shown in Table \ref{table:tracking2d_result}, where for clarity we just show the median values and variances. 

\setlength{\tabcolsep}{4pt}
\begin{table*}[!htbp]
\centering
\caption{2D Target Tracking Auto-tuning Results}
\begin{tabular}{*5c|*4c|*4c|*4c|c}
\toprule
& \multicolumn{4}{c}{\BO{} with $\Cnis$} & \multicolumn{4}{c}{\BO{} with  $\Jnis$} & \multicolumn{4}{c}{GPBO from \cite{chen2018weak}} & \multicolumn{4}{c}{Downhill Simplex from \cite{powell2002automated}}&
\multicolumn{1}{c}{GT}\\
& $v_0$ & $v_1$ & $w_0$ & $w_1$ & $v_0$ & $v_1$ & $w_0$ & $w_1$ & $v_0$ & $v_1$ & $w_0$ & $w_1$ & $v_0$ & $v_1$ & $w_0$ & $w_1$\\
\midrule
Med & \textbf{1.43} & \textbf{2.14} & \textbf{0.20} & \textbf{0.096}
& 1.61 & 1.41 & 0.14 & 0.16
& 3.377 & 3.378 & 0.150 & 0.152
&0.73 &0.65 &0.06 &0.08 & $v_0, v_1$ = 1, 2\\
Var & \textbf{0.24} & \textbf{0.43} & \textbf{0.002} & \textbf{3.2e-4}
&0.88 & 0.68 & 0.005 &0.02
&1.997 & 2.510 & 0.005 & 0.024
&0.33 &0.33 &0.002 &0.004 & $w_0,w_1$ = 0.2, 0.1\\ 
Mean & \textbf{1.49} & 2.40 & \textbf{0.21} & \textbf{0.098}
& 1.73 & \textbf{1.63} & 0.17 & 0.21
& 3.32 & 3.18 & 0.17 & 0.23
&0.72 &0.78 &0.06 &0.07 \\
\bottomrule
\end{tabular}
\label{table:tracking2d_result}
\end{table*}

\begin{figure*} [ht!]
\centering
\subfloat[NIS]{
\includegraphics[scale=0.95]{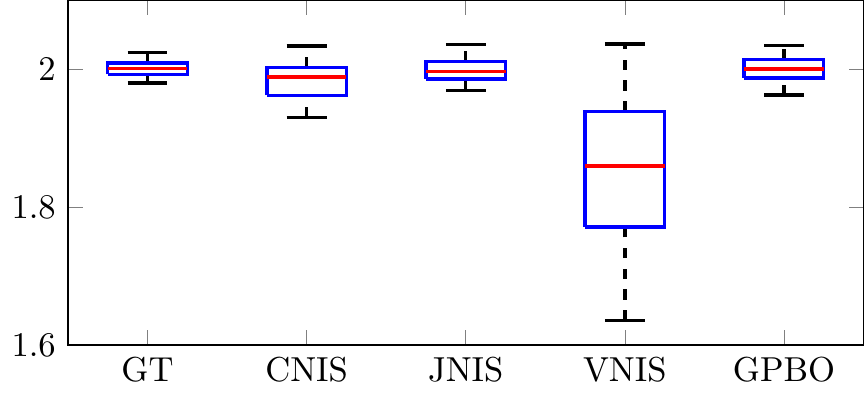}
\label{fig:nis_compare_dt01}
}
\subfloat[NEES]{
\includegraphics[scale=0.95]{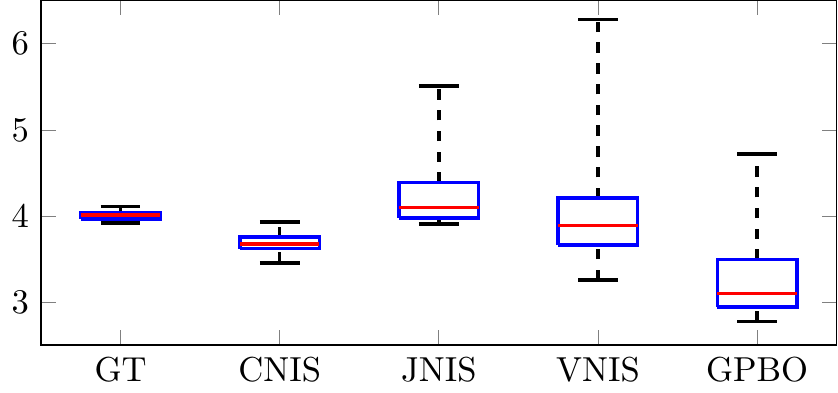}
\label{fig:nees_compare_dt01}
}
\\
\subfloat[NIS Variance]{
\includegraphics[scale=0.95]{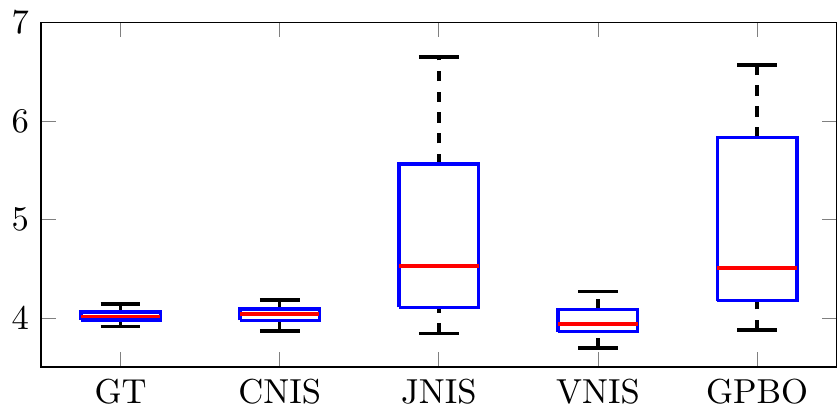}
\label{fig:nis_var_compare_dt01}
}
\subfloat[NEES Variance]{
\includegraphics[scale=0.95]{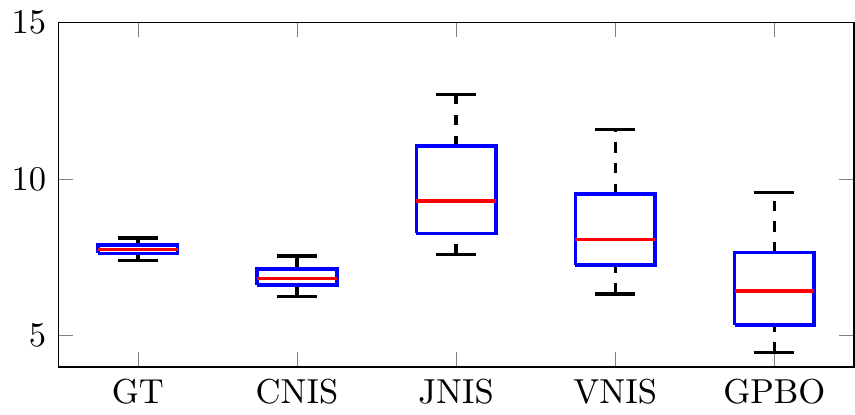}
\label{fig:nees_var_compare_dt01}
}\\
\caption{Statistical summaries over 50 optimization runs for $\E{\avgnis{k}}, \E{\avgnees{k}}, \E{\avgnis{k}\avgnis{k}}, \E{\avgnees{k}\avgnees{k}}$ when applying results of Kalman filter auto-tuning methods. 
CNIS has the lowest overall error and variation in the means and variances of the NIS and NEES with respect to GT, and therefore best maintains the expected $\chi^2$ distributions for consistency. }
\label{fig:nis_nees_chi_square_test}
\end{figure*}

For this problem, the degrees of freedom for the measurement innovation and the state error chi-square distributions are 2 and 4, respectively. Using eqs. \ref{eqn:nees_nis} and \ref{eqn:alpha_cov}, 
we see that a consistent filter must have
{\allowdisplaybreaks
\abovedisplayskip = 2pt
\abovedisplayshortskip = 2pt
\belowdisplayskip = 2pt
\belowdisplayshortskip = 2pt
\begin{align*}
    &\mathbb{E}[{\avgnees{k}}]{\approx} \ 4,\ \ \ \ \ \ \
    \mathbb{E}[{\avgnis{k}}]{\approx} \ 2,\\
    &\mathbb{E}[{\avgnees{k}\avgnees{k}}]{\approx} \ 8,\ \ 
    \mathbb{E}[{\avgnis{k}\avgnis{k}}]{\approx} \ 4.
\end{align*}
}
\noindent To validate whether these hold for TPBO, GPBO, and Downhill Simplex, we apply each method's tuning results to a Kalman Filter again with 120 time steps (with sample time $\Delta t = 0.1$ sec and ground truth values $[v_0, v_1, w_0, w_1] = [1,2,0.2,0.1]$) and record the statistics for $\E{\avgnis{k}}, \E{\avgnees{k}}, \E{\avgnis{k}\avgnis{k}}, \E{\avgnees{k}\avgnees{k}}$. 
The results 
across 50 such optimization runs are shown in Figure \ref{fig:nis_nees_chi_square_test} for TPBO with $\Cnis$, TPBO with $\Jnis$, and GPBO from \cite{chen2018weak} using $\Jnis$ with only one time discretization interval. Additionally, since $\Jnis$ only uses the mean term in $\Cnis$, we provide results where only the variance term $\savgnis$ in Eq. \eqref{eq:cnis} is used as a new cost function called $\Vnis$. 
%
The `GT' label in each plot denotes the distributions for NEES and NIS statistics produced by a Kalman filter across 50 runs.  
Note that NEES and NIS statistics were not consistent for the Downhill Simplex method and resulted in large standard deviations; since their scale makes it difficult to visualize the other results, the Downhill Simplex results are not included on the box plots.\\ 

In Figure \ref{fig:nis_compare_dt01},
we can see that TPBO and GPBO with $\Jnis$ have the most stable mean NIS value and most closely match the ground truth mean NIS value, followed by TPBO with $\Cnis$. This is not surprising because minimizing $\Jnis$ corresponds to matching the true mean NIS value. 
Although the mean NIS value from the TBPO $\Cnis$ optimization result is larger than results for $\Jnis$ from TBPO and GPBO, in reality, the difference is negligible. 
In contrast, $\Vnis$ results in a large mean NIS value because it only considers the variance term in $\Cnis$. 

Figure \ref{fig:nees_compare_dt01} similarly shows the resulting mean NEES values obtained by all methods. The TBPO $\Jnis$ method has the best median value but the TBPO $\Cnis$ method's NEES value is both stable and close to the expected value of 4. 
The VNIS and GPBO methods lead to high variance, and the GPBO median is heavily biased. 

The resulting NIS and NEES variances are shown in Figures \ref{fig:nis_var_compare_dt01} and \ref{fig:nees_var_compare_dt01}, where we can see that the TBPO $\Cnis$ method's NEES and NIS variances are again quite robust and close to the ideal expectation values, which means the $\chi^2$ consistency constraints are better maintained. 
$\Vnis$ gives a stable NIS variance value as expected, 
but leads to a less stable NEES variance. 
TBPO and GPBO with $\Jnis$ return biased and more volatile results for the NIS and NEES variances, thus violating the consistency constraints. 

\indent The key takeaway from these results is that auto-tuning via TPBO with the $\Cnis$ cost stably produces NIS and NEES variances that are close to their ideal expected values, and also stably produces correct expected NIS and NEES means. 
In this way, TPBO with $\Cnis$ maintains the $\chi^2$ consistency constraints and results in a filter with noise parameters that lead to uncertainty estimates accurately describing the actual estimation error statistics. On the other hand, the $\Jnis$ or $\Vnis$ cost functions for auto-tuning result in local minima that can only maintain either a stable NIS mean or NIS variance, instead of overall $\chi^2$ consistency (this phenomenon can further be seen in Sec. \ref{sct: sub_cascade_msp}).\\ 

Table \ref{table:tracking2d_result} quantitatively shows that the TBPO $\Cnis$ parameter optimization results are quite reasonable, and generally superior to those produced by the other methods.  However, when comparing the results to the 1D system, we can see the resulting $v_0$ mean and median for all method deviates noticeably from the ground truth value, resulting in relatively large variances for this parameter. 
For this 2D tracking problem, good filtering performance is still achievable even when one of the tuning variables $v_0,v_1,w_0,w_1$ is not close to the ground truth. In these cases, we still want to know if the resulting filter satisfies the filter consistency requirement. From Figure \ref{fig:nis_nees_chi_square_test}, we can observe whether the parameter optimization meets the $\chi^2$ constraints and thus if the resulting filter will remain consistent. 
The state estimation errors and $2\sigma$ bounds are plotted against time for the filter tuned via TPBO with $\Cnis$ are shown in
Figure \ref{fig:con_check_tracking2d}. Despite the fact that the tuned $v_0$ is biased from the ground truth, 
the resulting estimator performs well and indeed remains consistent. \\
%
%

\subsection{Cascade Mass-Spring-Damper System}
\label{sct: sub_cascade_msp}

\begin{figure}[t]
\centering
\includegraphics[width=0.38\textwidth]{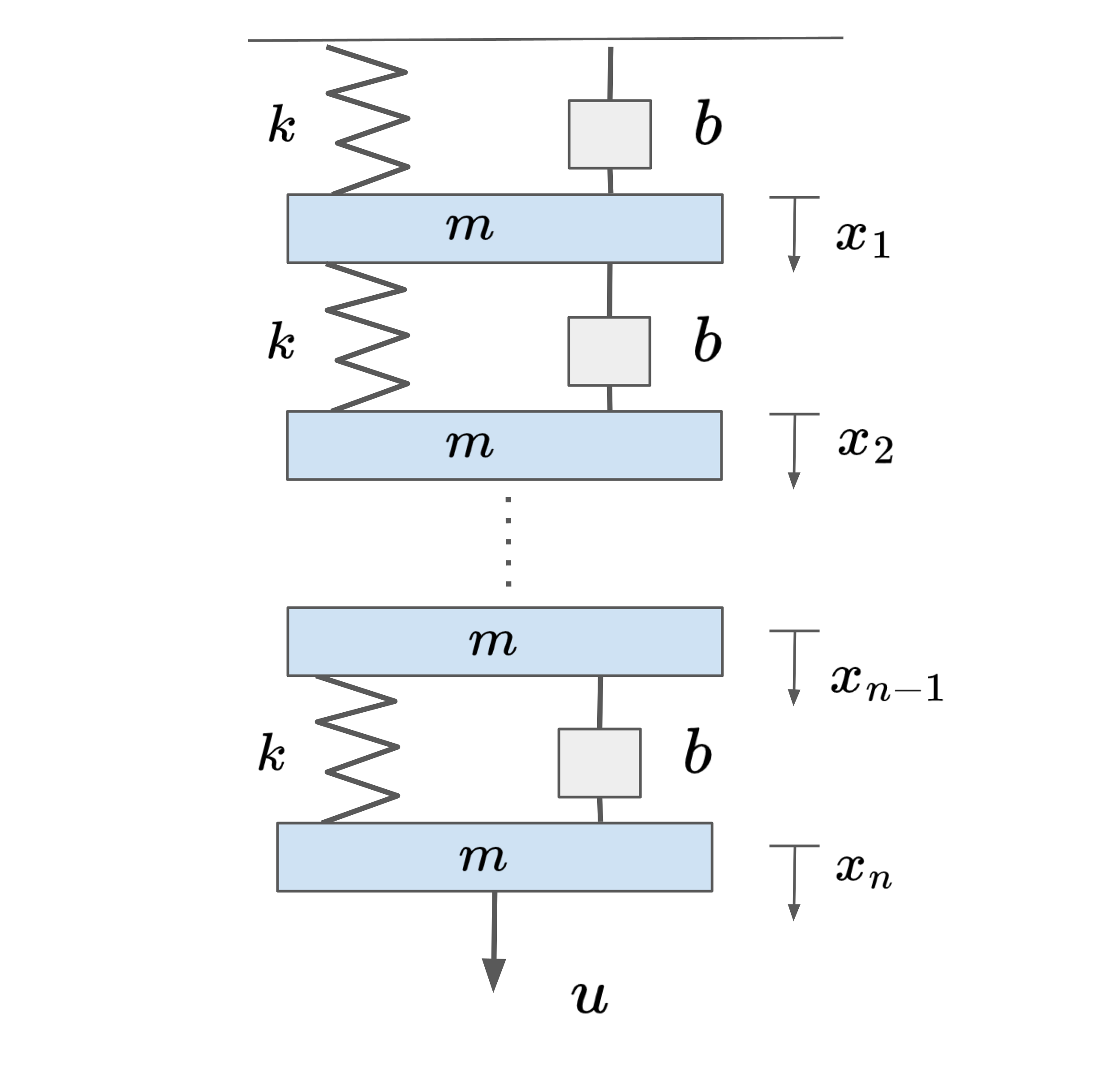}
\caption{Cascade mass-spring-damper system with $2n$ states. 
}
\label{fig:cascade_msp}
\end{figure}

\begin{figure*} [ht!]
\centering
\subfloat[NIS]{
\includegraphics[scale=0.95]{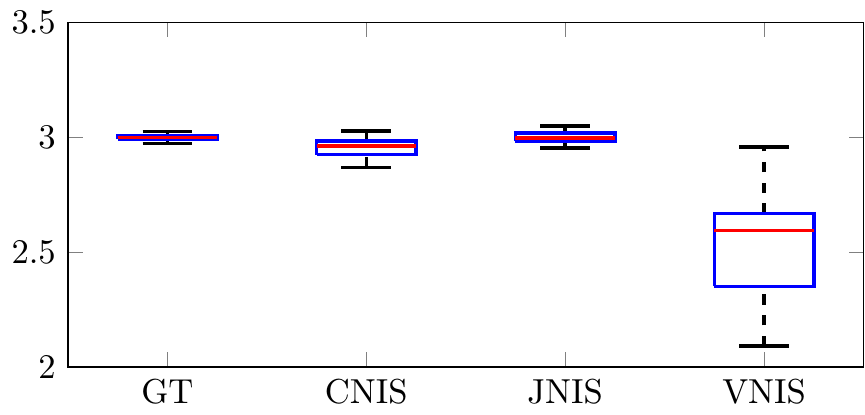}
\label{fig:cmsp_nis_compare_dt01}
}
\subfloat[NEES]{
\includegraphics[scale=0.95]{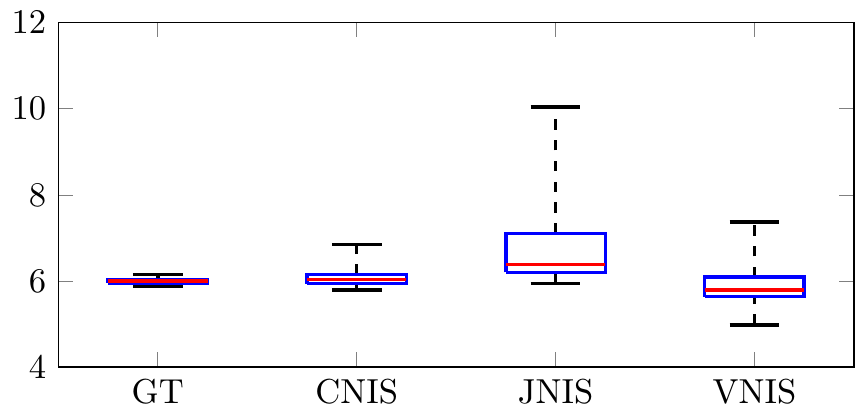}
\label{fig:cmsp_nees_compare_dt01}
}
\\
\subfloat[NIS Variance]{
\includegraphics[scale=0.95]{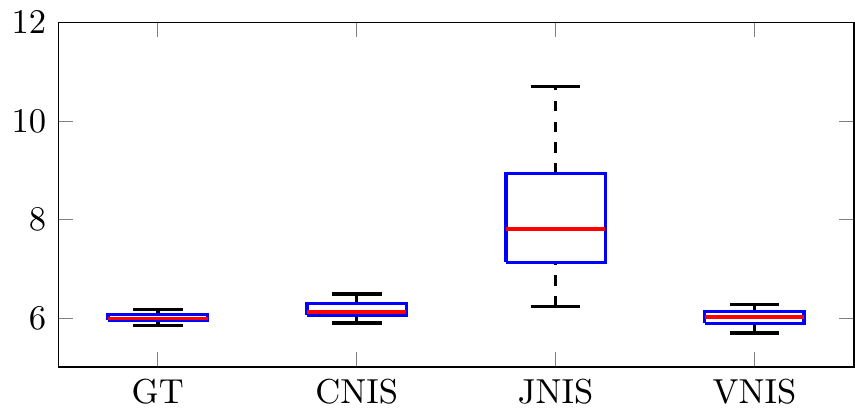}
\label{fig:cmsp_nis_var_compare_dt01}
}
\subfloat[NEES Variance]{
\includegraphics[scale=0.95]{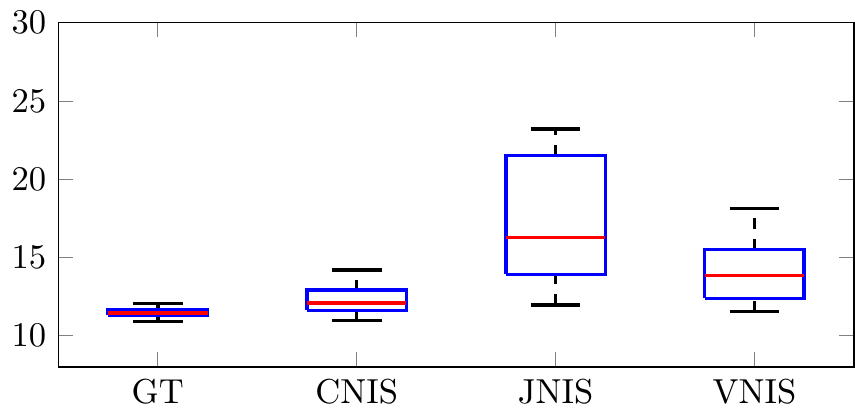}
\label{fig:cmsp_nees_var_compare_dt01}
}\\
\caption{ Similar to Figure \ref{fig:nis_nees_chi_square_test}, box plots of 50 independent optimizations are shown based on the various methods and then evaluate the resulting filters with optimized noise parameters. The $\Jnis$ and $\Vnis$ results' deviation from the ground truth shows the importance of considering both variance and mean term in $\Cnis$.} 
\label{fig:nis_nees_chi_square_test_6D}
\end{figure*}

A cascade of $n$ mass-spring-dampers 
is used to produce an even higher dimensional application. 
In the following test, $n=3$ and the state then is $\mathbf{x} = [ x_0,\dot{x}_0, x_1,\dot{x}_1,x_2,\dot{x}_2]$. 
The continuous-time model is 
{\allowdisplaybreaks
\abovedisplayskip = 2pt
\abovedisplayshortskip = 2pt
\belowdisplayskip = 2pt
\belowdisplayshortskip = 2pt
\begin{align}
&\mathbf{A} = \begin{bmatrix}
0 & 1 & 0 & 0 & 0 & 0  \\
-\frac{2k}{m} & -\frac{2b}{m} & \frac{k}{m} & \frac{b}{m} & 0 & 0 \\
0 & 0 & 0 & 1 & 0 & 0 \\
\frac{k}{m} & \frac{b}{m} & -\frac{2k}{m} & -\frac{2b}{m} & \frac{k}{m} & \frac{b}{m} \\
0 & 0 & 0 & 0 & 0 & 1 \\
0 & 0 & \frac{k}{m} & \frac{b}{m} & -\frac{k}{m} & -\frac{b}{m}
\end{bmatrix},\\
& \mathbf{G} = \begin{bmatrix} 0 & 0 & 0 & 0 & 0 & 1 \end{bmatrix} ^T, \\
&\mathbf{H} = \begin{bmatrix}
 1 & 0 & 0 & 0 & 0 & 0\\
 0 & 0 & 1 & 0 & 0 & 0\\
 0 & 0 & 0 & 0 & 1 & 0
\end{bmatrix},\\
&\boldsymbol{\Gamma} = \begin{bmatrix}
 0 & 1 & 0 & 0 & 0 & 0\\
 0 & 0 & 0 & 1 & 0 & 0\\
 0 & 0 & 0 & 0 & 0 & 1
\end{bmatrix}^T, \\
&\mathbf{V} = \begin{bmatrix}
 v_0 &  0   &  0\\
 0   & v_1  &  0\\
 0   &  0   & v_2
\end{bmatrix}, \ \ \
\mathbf{W} = \begin{bmatrix}
 w_0 &  0   &  0\\
 0   & w_1  &  0\\
 0   &  0   & w_2
\end{bmatrix}. 
\end{align}
}
The same control input as before is applied. 
We set the groundtruth $[v_0, v_1, v_2] = [1,2,3]$ and $[w_0, w_1,w_2] = [0.2, 0.1, 0.15]$. The initial sample number for Bayes optimization is 600 and the iteration count is 1200; the search range for all the process noise parameters is $v \in [0.1,5]$ and for the measurement noise parameters is $w \in [0.01, 1]$. We run 50 independent optimizations using the $\Cnis$, $\Jnis$, $\Vnis$ cost function for TPBO and GPBO with $\Jnis$. Similar to Figure \ref{fig:nis_nees_chi_square_test}, we run the filter with the 
optimized noise parameters and record $\E{\avgnis{k}}, \E{\avgnees{k}}, \E{\avgnis{k}\avgnis{k}}$, and $\E{\avgnees{k}\avgnees{k}}$, where a consistent filter should satisfy
{\allowdisplaybreaks
\abovedisplayskip = 2pt
\abovedisplayshortskip = 2pt
\belowdisplayskip = 2pt
\belowdisplayshortskip = 0pt
\begin{align*}
    &\E{\avgnees{k}}{\approx} \ 6, \ \ \ \ \ \ \ \ \
    \E{\avgnis{k}}{\approx} \ 3\\
    &\E{\avgnees{k}\avgnees{k}}{\approx} \ 12 \ \ \ \
    \E{\avgnis{k}\avgnis{k}}{\approx} \ 6.
\end{align*}
}
From Figure \ref{fig:nis_nees_chi_square_test_6D}, we observe that the $\Jnis$ cost function with \BO{} can only maintain the correct NIS mean but not the NIS variance, while the $\Vnis$ cost function for \BO{} can only maintain the correct NIS variance but not the NIS mean. The $\Cnis$ results are the closest to the ground truth in all four sub-plots. Also note that for this 6D optimization example, the deviation from the ground truth of the NIS variance is even larger for the $\Jnis$ results. This implies there might be more local minima when we implement a higher dimension system; this illustrates the significance of using $\Cnis$ to lock down the correct $\chi^2$ error distributions and tune the filter correctly. 
Note that the GPBO results using the method of \cite{chen2018weak} are ignored here since they produce quite large errors in the NIS and NEES variances. This effect underscores the importance of using multiple time discretization intervals for higher dimension systems. 
\\

\begin{figure*} [ht!]
\centering
\subfloat[$\Snu{\kCur|\kLst}(0,0)$]{
\includegraphics[scale=0.95]{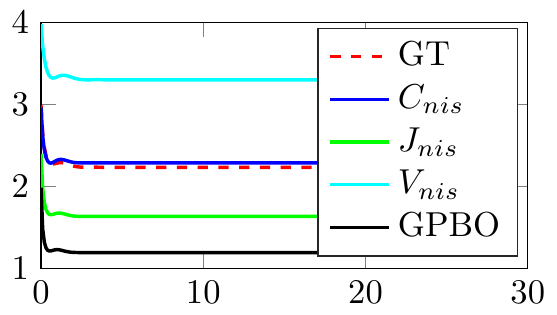}
\label{fig:sk0}
}
\subfloat[$\Snu{\kCur|\kLst}(1,1)$]{
\includegraphics[scale=0.95]{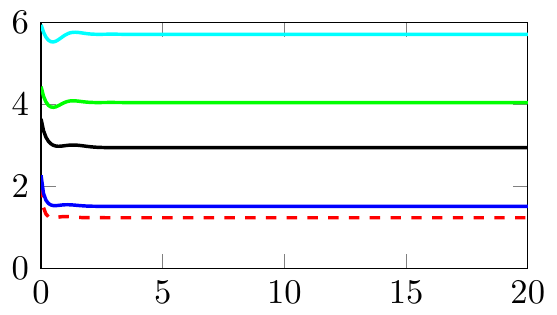}
\label{fig:sk1}
}
\subfloat[$\Snu{\kCur|\kLst}(2,2)$]{
\includegraphics[scale=0.95]{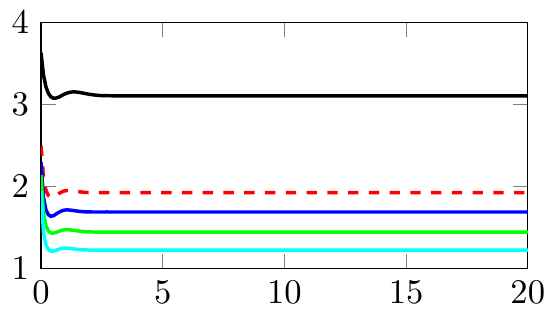}
\label{fig:sk2}
}\\
\caption{Innovation covariance matrix diagonal values over time for Kalman filters produced by each auto-tuning method vs. ground truth innovation covariance (GT, red dashed line).
}
\label{fig:innov_cov_diag}
\end{figure*}

Finally, Figure \ref{fig:innov_cov_diag}
shows the innovation covariance history $\Snu{\kCur|\kLst}$ for the Kalman filters produced by each auto-tuning method, where the innovation covariance for a correctly tuned filter should be close to the ground truth. 
We randomly choose one of the fifty optimization results for the $\Cnis,\Jnis,\Vnis$ and GPBO methods and run the resulting Kalman filters to record the innovation covariance history. 
For simplification, only the diagonal values of $\Snu{\kCur|\kLst}$ are plotted, corresponding to innovation variances of the sensed position states $\z = [x_0,x_1,x_2]$. 
These plots show that \BO{} using the multi-time discretization $\Cnis$ cost in eq. \eqref{eq: sum_cnis} produces results closest to the ground truth. 

\section{Conclusion}
\label{sct:conclusion}

%
This paper derived new cost metrics for Bayesian optimization auto-tuning of Kalman filter process noise and sensor noise covariance parameters, using either simulated or real data. Unlike previously developed cost metrics, the new cost metrics accurately account for both the mean and the variance of underlying state and measurement error statistics during the tuning process. When combined with Student's-t process surrogate regression models within Bayesian optimization, the new metrics improve the overall robustness of the parameter search, particularly when multiple time step discretizations are used for state space models that are converted from continuous time to discrete time. 
Three examples showed that the developed approach can scale to noise parameter searches in non-trivial high dimensional problems, converges toward optimal parameters faster and more stably than alternative black box optimization methods, and produces statistically consistent estimation results. 
The ability of our approach to reliably auto-tune Kalman filters using only recorded sensor data is a key capability in practical applications where obtaining ground truth data from truth model simulations or high-fidelity sensors is usually costly. 

\indent This work focused on auto-tuning of Kalman filters for linear systems to establish fundamentally useful insights that can be applied and developed further for a wide range of dynamic state estimation problems. Although linear estimators are still quite useful on their own in applications like target tracking, a logical direction for extending this work includes adaptation of the new metrics and \BO{} algorithm to auto-tuning of nonlinear estimators, e.g. based on Extended or Unscented Kalman filters \cite{Bar-Shalom2001, wan2000unscented} or batch algorithms, which are important for many real-world applications. 
For example, the NIS-based version of this work has already been applied to extrinsic parameter calibration in \cite{chen2018visual}, where the pose of a camera relative to other sensors mounted on a mobile robot was estimated using a batch processing algorithm. 
The effects of non-linearity and non-Gaussian error distributions lead to many challenging opportunities for exploring effective black box estimator auto-tuning strategies. 

\appendices

\section{Continuous to Discrete-Time Transformation}
\label{adx:cont_to_disc}

Although the stochastic linear systems we use are discrete time, these are derived from a continuous time formulation. 
The linear stochastic system which evolves according to the continuous-time model in eq. (\ref{eq:con_sys_model}) 
can be discretized to timesteps of length $\Delta{t}$ via \cite{simon2006optimal}
{\allowdisplaybreaks
\abovedisplayskip = 2pt
\abovedisplayshortskip = 2pt
\belowdisplayskip = 2pt
\belowdisplayshortskip = 2pt
\begin{equation}
\begin{aligned}
&\F{} = 
e^{\mathbf{A}\Delta t}, \ \
\mathbf{B} = 
\int_{0}^{\Delta t} e^{\mathbf{A} \tau} \mathrm{d} \tau, \ \
\R{} =  \frac{\W}{\Delta t},\\
&\Q{} =  \int_{0}^{\Delta t} e^{\mathbf{A}\tau} \mathbf{\Gamma} \V \mathbf{\Gamma}^T e^{\mathbf{A}^T\tau} \mathrm{d} \tau, 
\label{eq:integ}
\end{aligned}
\end{equation}
}
where van Loan's method provides a closed-form solution for $\Q{}$ in \eqref{eq:integ} \cite{BrownHwang2012}.


\section{TP Expected Improvement}
\label{adx:mvt_ei}

In this appendix, we expand the terms which appear in \eqref{eqn:EI_function}.

$u$ and $\sigma$ are the mean and variance of the conditional Student's-t distribution of $\mathbf{q}_{n+1}$, which is presented below in \eqref{con_MVT}.  $\Psi(\cdot)$ and $\psi(\cdot)$ are the CDF and PDF of the standard Student-t distribution $MVT_1(v, 0, 1)$. The conditional $MVT$ distribution is similar to the conditional multivariate Gaussian distribution: if we have $\mathbf{q}_{1:n+1}$ and $y(\mathbf{q}_{1:n+1})$ described by a multivariate $MVT$ pdf, then 
{\allowdisplaybreaks
\abovedisplayskip = 2pt
\abovedisplayshortskip = 2pt
\belowdisplayskip = 2pt
\belowdisplayshortskip = 2pt
\begin{align} \label{MVT_nplus1}
    \begin{split}
    \begin{bmatrix}
        y(\mathbf{q}_{1:n}) \\
        y(\mathbf{q}_{n+1})
    \end{bmatrix}
    & \sim 
    MVT_{n+1} \left(v+1,
    \begin{bmatrix}
        \Phi(\mathbf{q}_{1:n}) \\
        \Phi(\mathbf{q}_{n+1}) \\
    \end{bmatrix} \right|
    , \\
    & \left|\begin{bmatrix}
        K(\mathbf{q}_{1:n}, \mathbf{q}_{1:n}) & K(\mathbf{q}_{1:n}, \mathbf{q}_{n+1}) \\
        K(\mathbf{q}_{n+1}, \mathbf{q}_{1:n}) & K(\mathbf{q}_{n+1}, \mathbf{q}_{n+1}) \\
    \end{bmatrix} \right)
    \end{split}
\end{align}
}
where $K(\mathbf{q}_{1:n}, \mathbf{q}_{1:n})$ is the same as eq. (\ref{cov_matrix}), $K(\mathbf{q}_{1:n}, \mathbf{q}_{n+1}) = [k(\mathbf{q}_1 , \mathbf{q}_{n+1}), \cdots, k(\mathbf{q}_n , \mathbf{q}_{n+1})]^T$, and $K(\mathbf{q}_{n+1}, \mathbf{q}_{n+1}) = k(\mathbf{q}_{n+1} , \mathbf{q}_{n+1})$. 
\eqref{MVT_nplus1} can be written more simply as
{\allowdisplaybreaks
\abovedisplayskip = 2pt
\abovedisplayshortskip = 2pt
\belowdisplayskip = 2pt
\belowdisplayshortskip = 2pt
\begin{align} \label{sim_MVT_nplus1}
    \begin{split}
    \begin{bmatrix}
        y_1 \\
        y_2
    \end{bmatrix}
    \sim 
    MVT_{n+1}\left(v+1,
    \begin{bmatrix}
        \Phi_1 \\
        \Phi_2 \\
    \end{bmatrix}
    , 
    \begin{bmatrix}
        K_{11} & K_{12} \\
        K_{21} & K_{22} \\
    \end{bmatrix} \right)
    \end{split}
\end{align}
}
The conditional Student-t distribution of $y(\mathbf{q}_{n+1})$ is then given by \cite{ding2016conditional}
\begin{equation} \label{con_MVT}
    \begin{split}
        & y(\mathbf{q}_{n+1}|\mathbf{q}_{1:n}, y(\mathbf{q}_{1:n})) \sim MVT_1(v+n, u, \sigma) \\
        & u = \Phi_2 + K_{21}K_{11}^{-1}(y_1 - \Phi_1) \\
        & \sigma = \frac{v+d}{v+n} K_{22}-K_{21}K_{11}^{-1}K_{12} \\
        & d = (y_1 -  \Phi_1)^T K_{11}^{-1}(y_1 - \Phi_1)
    \end{split}
\end{equation}
Since the value of the prior mean function does not change the final result \cite{brochu2010tutorial}, we simplify the equations by setting $\Phi_1 = \mathbf{0}, \Phi_2 = 0$. 


\ifCLASSOPTIONcaptionsoff
  \newpage
\fi
\bibliographystyle{IEEEtran}
\bibliography{ekf_bayesopt.bib}

\newpage

\begin{IEEEbiography}[{\includegraphics[width=1in,height=1.25in,clip,keepaspectratio]{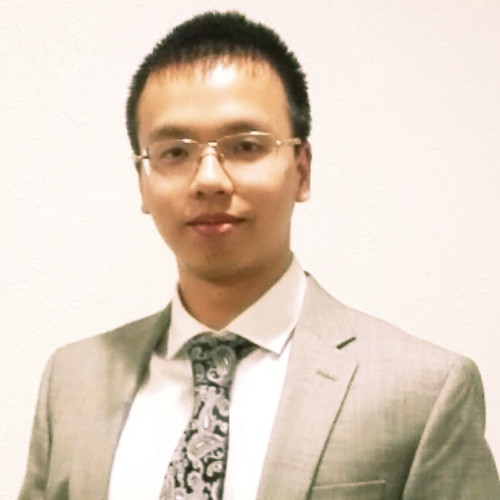}}]{Zhaozhong Chen} received the B.S. degree in Electrical Engineering in Tianjin University, China in 2016 and received his Ph.D. in University of Colorado Boulder, US in 2021 under the supervision of Prof. Christoffer Heckman. Zhaozhong's research interests focus on 3D reconstruction, visual odometry, multi-model navigation, neural rendering, state estimation and multi-sensor calibration.
\end{IEEEbiography}%

\begin{IEEEbiography}[{\includegraphics[width=1in,height=1.25in,clip,keepaspectratio]{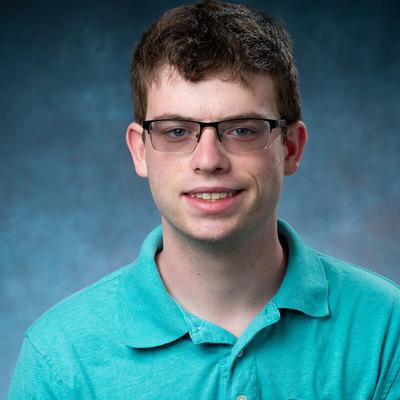}}]{Harel Biggie} received the B.S. degree in Electrical and Computer Engineering from the University of Rochester, Rochester, New York, USA, in 2018. He is currently working towards his Ph.D. in computer science with the Department of Computer Science, University of Colorado Boulder, Boulder, CO, USA. His research interests include natural language-based robot navigation, robotic exploration algorithms for unknown environments, and robust localization and mapping algorithms.
\end{IEEEbiography}

\begin{IEEEbiography}[{\includegraphics[width=1in,height=1.25in,clip,keepaspectratio]{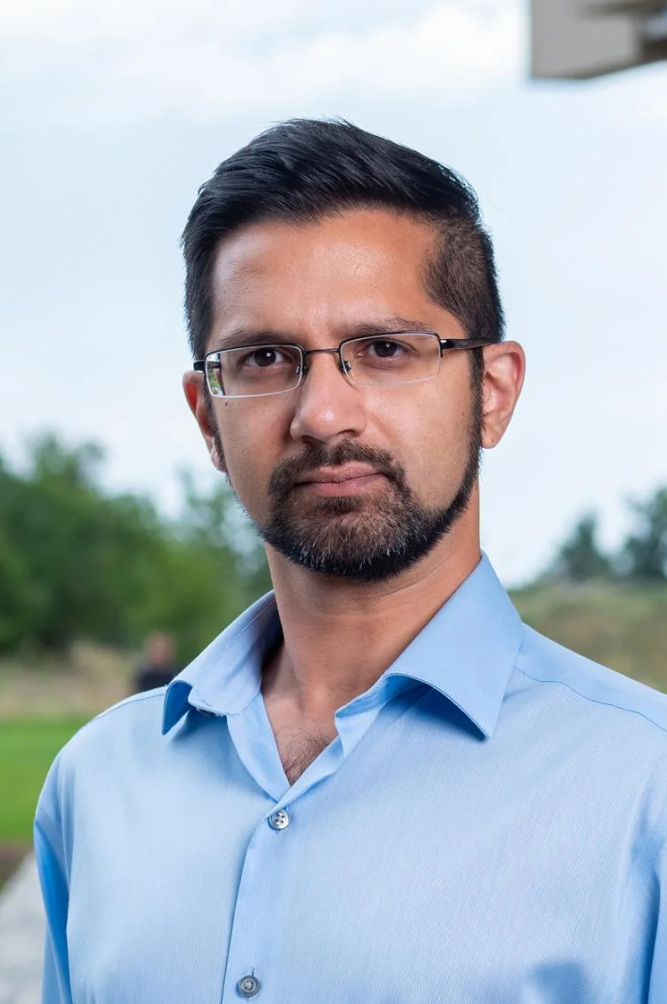}}]{Nisar Ahmed} (Member, IEEE), received the B.S.
degree in engineering from Cooper Union, New York,
NY, USA, in 2006, and the Ph.D. degree in mechanical engineering from Cornell University, Ithaca, NY,
USA, in 2012.
He is currently an Associate Professor and H.J.
Smead Faculty Fellow with the Smead Aerospace
Engineering Sciences Department, University of Colorado Boulder, Boulder, CO, USA. He directs the
Cooperative Human-Robot Intelligence (COHRINT)
Lab. His research interests include probabilistic modeling, estimation and control of autonomous systems, human-robot/machine
interaction, sensor fusion, and decision-making under uncertainty.
\end{IEEEbiography}

\begin{IEEEbiography}[{\includegraphics[width=1in,height=1.25in,clip,keepaspectratio]{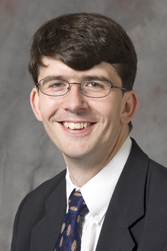}}]{Simon Julier} is a Professor of
Computer Science Department, University College
London (UCL), London, U.K. Before joining UCL,
he worked for nine years at the 3D Mixed and
Virtual Environments Laboratory, Naval Research
Laboratory, Washington, DC, where he was PI of the
Battlefield Augmented Reality System (BARS): a research effort to develop man-wearable systems
for providing situational awareness information. He
served as the Associate Director of the 3DMVEL
from 2005 to 2006. His research interests include user interfaces, distributed
data fusion, nonlinear estimation, and simultaneous localization and mapping.
\end{IEEEbiography}

\begin{IEEEbiography}[{\includegraphics[width=1in,height=1.25in,clip,keepaspectratio]{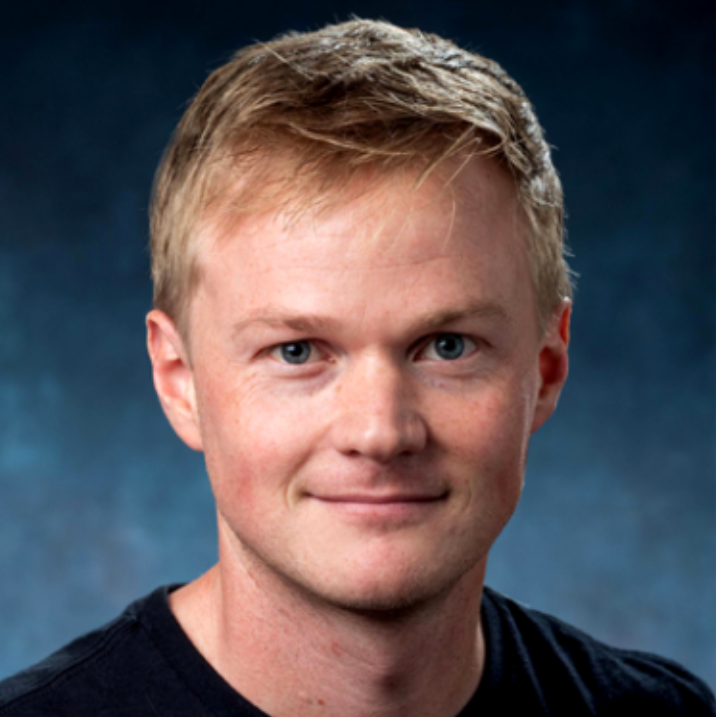}}]{Christoffer Heckman} (Senior Member, IEEE) is an Assistant Professor in the Department of Computer Science at the University of Colorado at Boulder and the Jacques I.\ Pankove Faculty Fellow in the College of Engineering and Applied Science. Prior to that, he was a postdoctoral researcher at the Naval Research Laboratory, in Washington, DC. He earned his BS in Mechanical Engineering from UC Berkeley in 2008 and his Ph.D. in Theoretical and Applied Mechanics from Cornell University in 2012. Heckman's research focuses on autonomy, perception, and field robotics.
\end{IEEEbiography}

\end{document}